\documentclass[conference]{IEEEtran}
\IEEEoverridecommandlockouts

\usepackage{cite}
\usepackage{amsmath,amssymb,amsfonts}
\usepackage{algorithmic}
\usepackage{graphicx}
\usepackage{textcomp}
\usepackage{xcolor}
\def\BibTeX{{\rm B\kern-.05em{\sc i\kern-.025em b}\kern-.08em
    T\kern-.1667em\lower.7ex\hbox{E}\kern-.125emX}}

\usepackage{circuitikz}
\usetikzlibrary{shapes.geometric, patterns, arrows.meta, shapes.arrows, fit, shadows.blur}
\usepackage{pgfplots,pgfplotstable}
\usepackage{subfig}
\usepackage{threeparttable}
\usepackage{multirow,hhline}
\usepackage{xcolor,colortbl}
\usepackage{todonotes}
\usepackage{setspace}

\makeatletter
\newcommand*\titleheader[1]{\gdef\@titleheader{#1}}
\AtBeginDocument{%
	\let\st@red@title\@title
	\def\@title{%
		\bgroup\normalfont\normalsize\centering\@titleheader\par\egroup
		\vskip0.5em\st@red@title}
}
\makeatother

\title{Efficient FPGA Implementation of Time-Domain Popcount for Low-Complexity Machine Learning \\
}
\titleheader{PREPRINT - Accepted at IEEE International Symposium on Asynchronous Circuits and Systems (ASYNC) 2025}

\begin{document}
\author{\IEEEauthorblockN{Shengyu Duan\IEEEauthorrefmark{1}, Marcos L. L. Sartori\IEEEauthorrefmark{1}, Rishad Shafik\IEEEauthorrefmark{1}, Alex Yakovlev\IEEEauthorrefmark{1}, Emre Ozer\IEEEauthorrefmark{2}} 
\IEEEauthorblockA{\IEEEauthorrefmark{1}Microsystems Research Group, Newcastle University, Newcastle upon Tyne, UK \ \ 
\IEEEauthorrefmark{2}Pragmatic Semiconductor, Cambridge, UK \\
\{shengyu.duan, marcos.sartori, rishad.shafik, alex.yakovlev\}.newcastle.ac.uk \ \ 
eozer@pragmaticsemi.com}
}

\maketitle
\thispagestyle{plain}
\pagestyle{plain}

\begin{abstract}
Population count (popcount) is a crucial operation for many low-complexity machine learning (ML) algorithms, including Tsetlin Machine (TM)-a promising new ML method, particularly well-suited for solving classification tasks. The inference mechanism in TM consists of propositional logic-based structures within each class, followed by a majority voting scheme, which makes the classification decision. In TM, the voters are the outputs of Boolean clauses. The voting mechanism comprises two operations: popcount for each class and determining the class with the maximum vote by means of an argmax operation.  

While TMs offer a lightweight ML alternative, their performance is often limited by the high computational cost of popcount and comparison required to produce the argmax result. In this paper, we propose an innovative approach to accelerate and optimize these operations by performing them in the time domain. Our time-domain implementation uses programmable delay lines (PDLs) and arbiters to efficiently manage these tasks through delay-based mechanisms. We also present an FPGA design flow for practical implementation of the time-domain popcount, addressing delay skew and ensuring that the behavior matches that of the model's intended functionality. By leveraging the natural compatibility of the proposed popcount with asynchronous architectures,
we demonstrate significant improvements in an asynchronous TM, including up to 38\% reduction in latency, 43.1\% reduction in dynamic power, and 15\% savings in resource utilization, compared to synchronous TMs using adder-based popcount. 
\end{abstract}


\section{Introduction}

There has been a shift towards low-complexity machine learning (ML) algorithms, offering competitive performance with fewer resources than deep neural networks. Tsetlin Machines (TMs) \cite{granmo2018tsetlin} and Binarized Neural Networks (BNNs) \cite{courbariaux2016binarized} are two prominent alternatives that leverage bit-wise operations, making them highly suitable for implementation on Field Programmable Gate Arrays (FPGAs). A TM is inherently logic-based, performing classification through propositional logic with Boolean inputs (Fig. \ref{fig:tm_overview} (a)). In contrast, a BNN represents an extreme case of quantized deep neural networks, encoding values with a single bit (+1/-1) and simplifying multiplications using XNOR operations (Fig. \ref{fig:tm_overview} (b)). 

\definecolor{mypink}{rgb}{0.858, 0.188, 0.478}

\newcommand{\overbar}[1]{\mkern 2.5mu\overline{\mkern-2.5mu#1\mkern-2.5mu}\mkern 2.5mu}

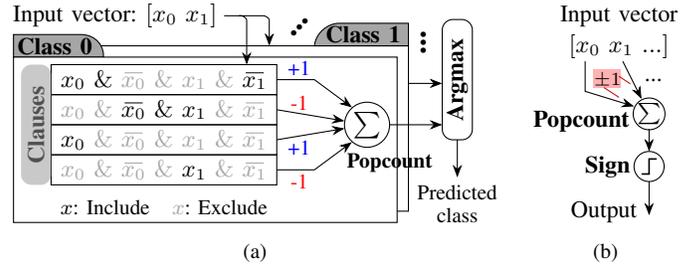
\begin{figure}[!htb]
\vspace{-0.4cm}
\centering
\subfloat[]{
	\begin{tikzpicture}[every node/.style={scale=1}, scale=1]
	
	\begin{scope}[xshift=0.15cm,yshift=0.15cm]
	
	\draw[-Stealth] (3,-0.2) -- (3.55,-0.2);
	
	\draw[fill=white] (-2,-1.9) rectangle ++ (5.1,2.2) node[] {};
	\draw[fill=gray!70!white] (1.9,0.6) -- (1.85,0.3) -- (3.1,0.3)   -- (3.1,0.6) [rounded corners=8pt] -- cycle;
	\node[] at (2.5,0.45) {\textbf{\small{Class 1}}};
	
	\node[] at (-0.8,0.7) {\small{Input vector: $[x_0 \ x_1]$}};
	
	\end{scope}
	
	\begin{scope}[xshift=0cm,yshift=0cm]
	
	\draw[fill=gray!70!white] 
	(-0.75,0.6) -- (-0.8,0.3) -- (-2,0.3)   -- (-2,0.6) [rounded corners=10pt] -- cycle;
	\draw[fill=white] (-2,-1.9) rectangle ++ (5.1,2.2) node[] {};
	\node[] at (-1.45,0.45) {\textbf{\small{Class 0}}};
	
	\draw
	(0, 0) node[text=gray!70!white] () {\small{$\textcolor{black}{x_0 \ \&} \ \overbar{x_0} \ \& \ x_1 \ \& \ \textcolor{black}{\overbar{x_1}}$}}
	;
	\draw[] (-1.5,-0.2) rectangle ++ (3,0.4) node[] {};
	
	\draw
	(0, -0.4) node[text=gray!70!white] () {\small{$x_0 \ \& \ \textcolor{black}{\overbar{x_0} \ \& \ x_1} \ \& \ \overbar{x_1}$}}
	;
	\draw[] (-1.5,-0.6) rectangle ++ (3,0.4) node[] {};
	
	\draw
	(0, -0.8) node[text=gray!70!white] () {\small{$\textcolor{black}{x_0} \ \& \ \overbar{x_0} \ \& \ x_1 \ \& \ \overbar{x_1}$}}
	;
	\draw[] (-1.5,-1) rectangle ++ (3,0.4) node[] {};
	
	\draw
	(0, -1.2) node[text=gray!70!white] () {\small{$x_0 \ \& \ \overbar{x_0} \ \& \ \textcolor{black}{x_1} \ \& \ \overbar{x_1}$}}
	;
	\draw[] (-1.5,-1.4) rectangle ++ (3,0.4) node[] {};
	
	\draw[draw=none, fill=gray!70!white, opacity=0.6, rounded corners] (-1.9,-1.4) rectangle ++(0.4,1.6) node[pos=.5, rotate=90] () {\textbf{\small{Clauses}}};
	
	\draw[-Stealth] (1.5,0) -- (2,0) -- (2.5,-0.37);
	\draw[-Stealth] (1.5,-0.4) -- (2.4,-0.55);
	\draw[-Stealth] (1.5,-0.8) -- (2.4,-0.65);
	\draw[-Stealth] (1.5,-1.2) -- (2,-1.2) -- (2.5,-0.83);
	
	\node[] at (1.8,0.15) {\textcolor{blue}{\footnotesize{+1}}};
	\node[] at (1.8,-0.3) {\textcolor{red}{\footnotesize{-1}}};
	\node[] at (1.8,-0.9) {\textcolor{blue}{\footnotesize{+1}}};
	\node[] at (1.8,-1.35) {\textcolor{red}{\footnotesize{-1}}};
	\node[draw,circle,minimum size=0.6cm,inner sep=0pt] at (2.7,-0.6) {$\sum$};
	\draw[-Stealth] (3,-0.6) -- (3.7,-0.6);
	
	\node[] at (3,-1.1) {\textbf{\footnotesize{Popcount}}};
	
	\node[] at (0,-1.7) {\footnotesize{$x$: Include \ \ \textcolor{gray!70!white}{$x$}: Exclude}};
	
	\draw[]
	(0.8,0.85) -- (1.3,0.85)
	(1,0.85) to [bend left] (1.1,0.8)
	(1.3,0.85) to [bend left] (1.4,0.8)
	;
	
	\draw[-Stealth] (1.1,0.8) -- (1.1,0.2);
	
	\draw[-Stealth] (1.4,0.8) -- (1.4,0.45);
	
	\node[rotate=45] at (1.8,0.7) {\large{\textbf{...}}};
	
	\draw[rounded corners] (3.7,-0.8) rectangle ++(0.4,1.6) node[pos=.5, rotate=90] () {\textbf{\small{Argmax}}};
	\node[rotate=90] at (3.45,0.5) {\large{\textbf{...}}};
	
	\draw[-Stealth] (3.9,-0.8) -- (3.9,-1.3);
	\node[] at (3.9, -1.5) {\footnotesize{Predicted}};
	\node[] at (3.9, -1.8) {\footnotesize{class}};
	
	\end{scope}
	
	\end{tikzpicture}
}
\subfloat[]{
\begin{tikzpicture}[every node/.style={scale=1}, scale=1]

\node[] at (-0.4,0) {\small{Input vector}};
\node[] at(-0.4,-0.4)  {\small{$[x_0 \ x_1 \ ...]$}};
\draw[-Stealth] (-0.85,-0.6) -- (-0.85,-1) -- (-0.2,-1.2);
\draw[-Stealth] (-0.35,-0.6) -- (-0.1,-1.1);
\node[] at (0.06,-0.85) {\small{...}};
\draw[draw=none, fill=red!50!white, opacity=0.6] (-0.75,-0.95) rectangle ++ (0.37,0.25) node[] {};
\node[] at (-0.55,-0.85) {\scriptsize{$\pm 1$}};
\draw[red!75!black, very thin] 
(-0.6,-0.95) -- (-0.4,-1.08)
(-0.4,-0.8) -- (-0.25,-0.9)
;
\node[draw,circle,minimum size=0.4cm,inner sep=0pt] at (0,-1.3) {\footnotesize{$\sum$}};
\node[] at (-0.9,-1.4) {\small{\textbf{Popcount}}};
\draw[-Stealth] (0,-1.5) -- (0,-1.8);
\node[draw,circle,minimum size=0.4cm,inner sep=0pt] at (0,-2) {};
\draw[semithick] (-0.1,-2.1) -- (0,-2.1) -- (0,-1.9) -- (0.1,-1.9);
\node[] at (-0.55,-2) {\small{\textbf{Sign}}};
\draw[-Stealth] (0,-2.2) -- (0,-2.65);
\node[] at (-0.6,-2.6) {\small{Output}};

\end{tikzpicture}
}

\caption{(a) TMs, where each TM is assigned to a certain class and each clause has been trained to recognize a pattern of Boolean inputs, represented by propositional logic. Popcount counts the number of clauses supporting (+1) and opposing (-1) each class, with the classification determined by the class with the highest popcount, using argmax. (b) A BNN neuron. $x_0$ and $x_1$ are input features or activation values.}
\label{fig:tm_overview}
\vspace{-0.5cm}
\end{figure}

Population count (otherwise known as popcount) and comparison (argmax) are critical operations in both TMs and BNNs. In a TM, these operations function as a majority vote mechanism to determine classification outcomes (Fig. \ref{fig:tm_overview} (a)). On the other hand, in a BNN, popcount serves as the accumulation function for each neuron, followed by a comparator that applies the sign function by comparing the result to zero for activation (Fig. \ref{fig:tm_overview} (b)). However, studies have identified popcount and comparison as bottlenecks in BNN implementations, increasing latency and resource consumption due to their relatively low hardware efficiency compared to logic operations \cite{anderson2019photonic, tanigawa2024efficient}. For the first time, we will present these operations as bottlenecks for TMs in Section \ref{sec:case}.

To overcome this bottleneck, some efforts have been dedicated to developing cost-efficient and high-performance popcount accelerators and compressors, with a primary focus on adder-based architectures \cite{tanigawa2024efficient, li2022fpga, kim2018fpga, ma2023fpga}. In this paper, we introduce a paradigm shift by transitioning popcount and argmax operations to the time domain. For ML algorithms including TMs and BNNs, the core functionality remains unaffected by this transformation, as their outputs are typically determined by relative magnitudes rather than absolute values.

The basic principle of our method is as follows: the higher the popcount the smaller the delay of the corresponding delay-line. This combination of time-domain functionality naturally suits the use of asynchronous logic design. To implement this method in FPGA, we use programmable delay lines (PDLs), constructed using Lookup Tables (LUTs), to function as population counters (pop counters), while arbiters are employed as comparators that respond based on signal arrival times.
This operation's output is one-hot, and computations are interleaved with spacers, making the use of completion detectors the natural choice.
Furthermore, we propose a design flow for placement, pin assignment, and routing of PDLs to mitigate delay skew introduced by generic FPGA implementations.

We leverage this advantage to design asynchronous circuit for the TM inference using MOUSETRAP \cite{singh2007mousetrap} on a Xilinx Zynq XC7Z020 FPGA (PYNQ-Z1), using a single-rail bundled datapath and two-phase handshake protocol. 
Our case study demonstrates a low latency and energy-efficient inference process for asynchronous TMs, achieved through the implementation of time-domain popcount. This results in enhanced throughput (up to 38\% lower latency), reduced power consumption (up to 43.1\% less power), and lower resource utilization (up to 15\% less) compared to conventional adder-based popcount implementations, particularly for multi-class classification tasks that involve comparisons across many entities.

We make the following key \textit{\textbf{contributions}}:
\begin{itemize}
\item A novel time-domain circuit design for popcount and comparison, enabling cost-efficient and scalable implementations for TMs and possibly BNNs.
\item A FPGA design flow for implementing time-domain popcount, addressing propagation delay uncertainties to ensure reliable and efficient performance.
\item High-performance asynchronous architectures for TMs with time-domain pop counters and comparators, significantly improving throughput and power efficiency.
\end{itemize}

The paper is organized as follows: Section \ref{sec:related} reviews related work on popcount circuits and PDLs. Section \ref{sec:time-domain} presents our time-domain popcount and comparison design and FPGA implementation flow. Section \ref{sec:case} showcases an asynchronous TM case study, compares it with existing approaches. Section \ref{sec:conc} concludes the paper and outlines future work.

\section{Related Work} \label{sec:related}
We overview state-of-the-art digital popcount designs and PDLs, focusing on their suitability for FPGA implementation.

\subsection{Popcount} \label{sec:popcount}
Conventional popcount designs primarily rely on binary full adder trees to sum input bits. Recent research has largely focused on optimizing wide adders, while efforts to accelerate or compress popcount adders remain limited. One improvement is an 8-bit popcount design that reduces resource usage \cite{dalalah2006new}. However, for longer input vectors, this approach requires additional levels to aggregate multiple 8-bit popcount results, ultimately leading to a tree-based adder architecture.

A more recent approach leverages 6-input LUTs in modern FPGAs to compress popcount trees, where three LUTs collectively function as a 6-bit popcounter, producing a 3-bit output \cite{liang2018fp}. Another design, optimized based on ripple carry adders, introduces an additional chain to propagate the sum of each full adder \cite{kim2018fpga}. While this method achieves modest resource savings, it increases latency compared to conventional popcount trees. Further optimizations based on these works have been proposed by sharing logic elements \cite{li2022fpga, ma2023fpga}.

Most existing adder-based approaches remain within a similar design space, where improving one metric typically comes at the expense of others. More importantly, the comparison of multiple popcount results—an essential operation in applications such as TMs and the output layer of BNNs—introduces significant overhead in terms of latency and resource consumption when using digital comparators \cite{pedroni2004compact}. In ML tasks involving a large number of classes, this comparison (argmax) becomes a major bottleneck. 

\subsection{Programmable Delay Line (PDL)}
PDLs have been practically implemented on FPGAs in previous works, utilizing a cascade of programmable delay elements, where each element consists of a single LUT that buffers or inverts an input signal with its delay multiplexed by other inputs \cite{majzoobi2010fpga, mahalat2019efficient, anandakumar2022implementation, sahoo2015towards}. These PDLs are commonly used for arbiter physical unclonable functions (PUFs), which compare signals racing through two symmetric PDLs and generate responses based on cumulative delays of all units in each path.

However, PDLs originally for arbiter PUFs cannot be directly applied for time-domain popcount. First, PUF outputs are determined by specific input vector, whereas popcount outputs depend on input Hamming weight; for example, the output should remain the same for input vectors ``0...01" and ``10...0" in popcount, which is not the case for PUFs. Second, while PUFs exploit intrinsic process variations to generate device-specific responses, popcount must mitigate these variations for consistency and device-independency.

A recent study proposed using PDLs specifically for TM popcount followed by asynchronous arbitration for argmax \cite{lan2024asynchronous}. However, this work lacks validation through physical implementation of integrated circuit or FPGA. We emphasize that achieving accurate popcount using PDLs requires both structural and physical uniformity, the latter of which can only be ensured through careful physical design. On FPGAs, routing delays dominate over logic delays, 
necessitating precise placement and routing to maintain an unskewed relationship between input Hamming weight and popcount. This effort is crucial for the deployment of PDLs for popcount.

\section{Time-domain Popcount on FPGA} \label{sec:time-domain}
\subsection{Time-domain Popcount and Comparison} \label{sec:circuit}
We present the overall architecture for the time-domain popcount and comparison in Fig. \ref{fig:pdl}.

\definecolor{mypink}{rgb}{0.858, 0.188, 0.478}

\tikzset{
	multiplexer/.style={
		draw,
		trapezium,
		shape border uses incircle, 
		shape border rotate=270,
		minimum size=18pt,
		fill=white
	}  
}

\tikzset{flipflop myD1/.style={flipflop,
		flipflop def={t2=\scalebox{1}[1.67]{D}, t5=\scalebox{1}[1.67]{Q}, cd=1, font=\Large}}
}
\tikzset{flipflop myD2/.style={flipflop,
		flipflop def={t2=\scalebox{1}[1.67]{D}, t5=\scalebox{1}[1.67]{Q}, cu=1, font=\Large}}
}

\pgfplotstableread{
	A	delay
	1	1013
	2	811
	3	648
	4	774
	5 	414
	6	355
}\PinAssign

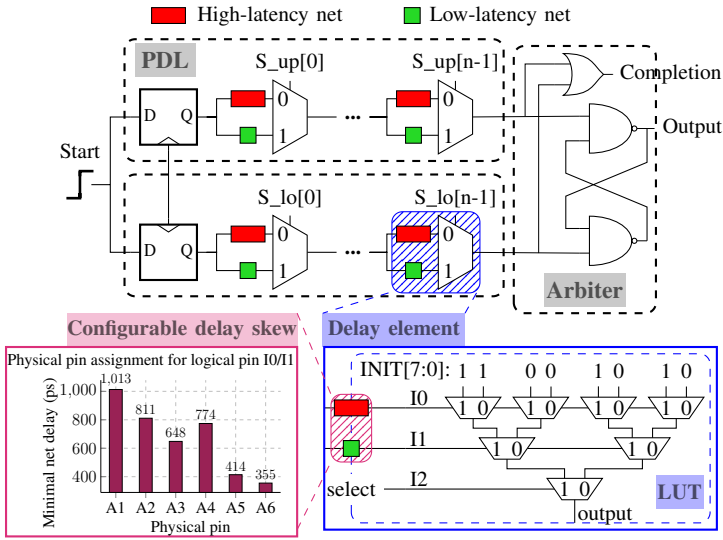
\begin{figure}[!htb]
\hspace{-0.9cm}
\centering
\begin{circuitikz}[font=\small, scale=0.9, every node/.style={scale=0.9}]

\node [draw, shape=rectangle, minimum width=0.5cm, minimum height=0.2cm, fill=red, anchor=center] at (0, 1.5) {};
\node [] at (1.5, 1.5) {High-latency net};
\node [draw, shape=rectangle, minimum width=0.2cm, minimum height=0.2cm, fill=green!70!gray, anchor=center] at (3.6, 1.5) {};
\node [] at (4.9, 1.5) {Low-latency net};

\ctikzset{multipoles/flipflop/clock wedge size=0.4}

\draw (0,0) node[flipflop myD1, scale=0.5, yscale=0.65](D1){}
(D1.pin 5) node[right=1cm] (de0_u_pos){}

(de0_u_pos) node[multiplexer, scale=0.8](de0_u){}
(de0_u_pos) node[yshift=0.27cm,xshift=-0.1cm] (de0_u0) {0}
(de0_u_pos) node[yshift=-0.27cm,xshift=-0.1cm] (de0_u1) {1}
(de0_u_pos) node[yshift=0.8cm] (de0_sel) {S\_up[0]}
(de0_u_pos) node[yshift=0.27cm, xshift=-0.12cm] (de0_pin0) {}
(de0_u_pos) node[yshift=-0.27cm, xshift=-0.12cm] (de0_pin1) {}
(de0_u_pos) node[yshift=0.27cm] (de0_pins) {}
(de0_u_pos) node[xshift=0.13cm] (de0_pino) {}
(D1.pin 5) --++ (0.15cm,0) |- (de0_pin0) 
(D1.pin 5) --++ (0.15cm,0) |- (de0_pin1)
(de0_sel) -- (de0_pins)
(de0_pin0) node[xshift=-0.5cm] (de0_slow) {}
(de0_pin1) node[xshift=-0.5cm] (de0_fast) {}
(de0_pino) node[xshift=0.8cm] (de0_uout) {\textbf{...}}
(de0_pino) -- (de0_uout)
;
\node [draw, shape=rectangle, minimum width=0.5cm, minimum height=0.2cm, fill=red, anchor=center] at (de0_slow) {};
\node [draw, shape=rectangle, minimum width=0.2cm, minimum height=0.2cm, fill=green!70!gray, anchor=center] at (de0_fast) {};

\draw
(de0_uout) node[right=1.25cm] (den_u_pos){}
(den_u_pos) node[multiplexer, scale=0.8](den_u){}
(den_u_pos) node[yshift=0.27cm,xshift=-0.1cm] (den_u0) {0}
(den_u_pos) node[yshift=-0.27cm,xshift=-0.1cm] (den_u1) {1}
(den_u_pos) node[yshift=0.8cm] (den_sel) {S\_up[n-1]}
(den_u_pos) node[yshift=0.27cm, xshift=-0.12cm] (den_pin0) {}
(den_u_pos) node[yshift=-0.27cm, xshift=-0.12cm] (den_pin1) {}
(den_u_pos) node[yshift=0.27cm] (den_pins) {}
(den_u_pos) node[xshift=0.13cm] (den_pino) {}
(de0_uout) --++ (0.5cm,0) |- (den_pin0) 
(de0_uout) --++ (0.5cm,0) |- (den_pin1)
(den_sel) -- (den_pins)
(den_pin0) node[xshift=-0.5cm] (den_slow) {}
(den_pin1) node[xshift=-0.5cm] (den_fast) {}
;
\node [draw, shape=rectangle, minimum width=0.5cm, minimum height=0.2cm, fill=red, anchor=center] at (den_slow) {};
\node [draw, shape=rectangle, minimum width=0.2cm, minimum height=0.2cm, fill=green!70!gray, anchor=center] at (den_fast) {};

\draw[thick, dashed, rounded corners] (-0.65,-0.65) rectangle ++ (5.6,1.8) node[] {};
\draw[draw=none, fill=gray!70!white, opacity=0.6] (-0.5,0.6) rectangle ++(0.9,0.5) node[pos=.5] () {\textbf{\normalsize{PDL}}};

\draw [line width=0.2pt]
(den_pino) node[xshift=2.6cm, yshift=-0.17cm] (arbiter_u_pos){}
(arbiter_u_pos) node[nand port, scale=0.65] (arbiter_u) {}
(den_pino) -- (arbiter_u.in 1)
(arbiter_u.out) node[right=0.1cm] (result) {Output}
(arbiter_u.out) -- (result)
(arbiter_u.in 1) node[xshift=-0.8cm,yshift=-0.12cm] (completion_u) {}
; 

\begin{scope}[xshift=0cm,yshift=-2cm]
	
\draw (0,0) node[flipflop myD2, scale=0.5, yscale=0.65](D2){}
(D2.pin 5) node[right=1cm] (de0_l_pos){}

(de0_l_pos) node[multiplexer, scale=0.8](de0_l){}
(de0_l_pos) node[yshift=0.27cm,xshift=-0.1cm] (de0_l0) {0}
(de0_l_pos) node[yshift=-0.27cm,xshift=-0.1cm] (de0_l1) {1}
(de0_l_pos) node[yshift=0.8cm] (de0_lsel) {S\_lo[0]}
(de0_l_pos) node[yshift=0.27cm, xshift=-0.12cm] (de0_lpin0) {}
(de0_l_pos) node[yshift=-0.27cm, xshift=-0.12cm] (de0_lpin1) {}
(de0_l_pos) node[yshift=0.27cm] (de0_lpins) {}
(de0_l_pos) node[xshift=0.13cm] (de0_lpino) {}
(D2.pin 5) --++ (0.15cm,0) |- (de0_lpin0) 
(D2.pin 5) --++ (0.15cm,0) |- (de0_lpin1)
(de0_lsel) -- (de0_lpins)
(de0_lpin0) node[xshift=-0.5cm] (de0_lslow) {}
(de0_lpin1) node[xshift=-0.5cm] (de0_lfast) {}
(de0_lpino) node[xshift=0.8cm] (de0_lout) {\textbf{...}}
(de0_lpino) -- (de0_lout)
;
\node [draw, shape=rectangle, minimum width=0.5cm, minimum height=0.2cm, fill=red, anchor=center] at (de0_lslow) {};
\node [draw, shape=rectangle, minimum width=0.2cm, minimum height=0.2cm, fill=green!70!gray, anchor=center] at (de0_lfast) {};

\draw[rounded corners, blue, pattern=north east lines, pattern color=blue] (3.3,-0.6) rectangle ++ (1.3,1.2) node[] {};

\draw
(de0_lout) node[right=1.25cm] (den_l_pos){}
(den_l_pos) node[multiplexer, scale=0.8](den_l){}
(den_l_pos) node[yshift=0.27cm,xshift=-0.1cm] (den_l0) {0}
(den_l_pos) node[yshift=-0.27cm,xshift=-0.1cm] (den_l1) {1}
(den_l_pos) node[yshift=0.8cm] (den_lsel) {S\_lo[n-1]}
(den_l_pos) node[yshift=0.27cm, xshift=-0.12cm] (den_lpin0) {}
(den_l_pos) node[yshift=-0.27cm, xshift=-0.12cm] (den_lpin1) {}
(den_l_pos) node[yshift=0.27cm] (den_lpins) {}
(den_l_pos) node[xshift=0.13cm] (den_lpino) {}
(de0_lout) --++ (0.5cm,0) |- (den_lpin0) 
(de0_lout) --++ (0.5cm,0) |- (den_lpin1)
(den_lsel) -- (den_lpins)
(den_lpin0) node[xshift=-0.5cm] (den_lslow) {}
(den_lpin1) node[xshift=-0.5cm] (den_lfast) {}
;
\node [draw, shape=rectangle, minimum width=0.5cm, minimum height=0.2cm, fill=red, anchor=center] at (den_lslow) {};
\node [draw, shape=rectangle, minimum width=0.2cm, minimum height=0.2cm, fill=green!70!gray, anchor=center] at (den_lfast) {};

\draw[thick, dashed, rounded corners] (-0.65,-0.65) rectangle ++ (5.6,1.8) node[] {};

\draw[draw=none, fill=blue!50!white, opacity=0.6] (2.28,-1.4) rectangle ++(2.1,0.5) node[pos=.5] () {\textbf{Delay element}};
\draw[dashed, blue] 
(3.25, -0.6) -- (2.28,-0.9)
(4.55, -0.6) -- (4.38,-0.9)
;  

\draw [line width=0.2pt]
(den_lpino) node[xshift=2.6cm, yshift=0.17cm] (arbiter_l_pos){}
(arbiter_l_pos) node[nand port, scale=0.65] (arbiter_l) {}
(den_lpino) -- (arbiter_l.in 2)
(arbiter_l.in 2) node[xshift=-0.6cm,yshift=-0.12cm] (completion_l) {}
;

\draw[thick, dashed, rounded corners] (5.1,-0.9) rectangle ++ (2.1,3.95) node[] {};
\draw[draw=none, fill=gray!70!white, opacity=0.6] (5.55,-0.8) rectangle ++(1.2,0.5) node[pos=.5] () {\textbf{\normalsize{Arbiter}}}; 

\end{scope}

\node (mid) at ($(D1.pin 2)!0.5!(D2.pin 2)$) {}; 
\draw 
(D1.down) -- (D2.up)
(arbiter_u.out) --++ (0,-0.5cm) --++ (-1.2cm,-0.5cm) |- (arbiter_l.in 1)
(arbiter_l.out) --++ (0,0.5cm) --++ (-1.2cm,0.5cm) |- (arbiter_u.in 2)
(D1.pin 2) --++ (-0.3cm,0) |- (D2.pin 2)
(mid) node[left=0.05cm] (mid_left) {}
(mid_left) --++ (-0.4cm, 0)
(mid_left) node[left=0.4cm] (trig) {}
; 
\draw[line width=1pt]
(trig) --++(0,0.2cm) --++(0.2cm,0)
(trig) --++(0,-0.2cm) --++(-0.2cm,0)
(trig) --++(0,-0.2cm) --++(0,0.4cm)
;
\node[yshift=0.5cm] at (trig) {Start};
\draw [line width=0.2pt]
(arbiter_u) node[yshift=0.8cm,xshift=-0.5cm] (completion_pos){}
(completion_pos) node[or port, scale=0.55] (completion) {}
(completion_l) |- (completion.in 2)
(completion_u) |- (completion.in 1)
(completion.out) -- 
(completion.out) node[right=0cm] (comp) {Completion}
(completion.out) -- (comp)
;

\begin{scope}[xshift=4.5cm,yshift=-4.3cm]
\draw
(-1,0) -- (3,0)
(-0.8,0) node[yshift=0.15cm] {I0} 

(0,0) node[multiplexer, xscale=0.6, yscale=0.5, rotate=-90](mux1){}
(0,0) node[xshift=0.15cm] (mux1_0) {0}
(0,0) node[xshift=-0.15cm] (mux1_1) {1}
(0,0) node[yshift=0.02cm, xshift=0.15cm] (mux1_pin0) {}
(0,0) node[yshift=0.02cm, xshift=-0.15cm] (mux1_pin1) {}
(mux1_1) node[yshift=0.55cm] (init7) {1}
(mux1_0) node[yshift=0.55cm] (init6) {1}
(init7) node[xshift=-0.85cm] () {INIT[7:0]:}
(init7) -- (mux1_pin1)
(init6) -- (mux1_pin0)

(mux1) node[xshift=1cm] (mux2_pos) {} 
(mux2_pos) node[multiplexer, xscale=0.6, yscale=0.5, rotate=-90](mux2){}
(mux2_pos) node[xshift=0.15cm] (mux2_0) {0}
(mux2_pos) node[xshift=-0.15cm] (mux2_1) {1}
(mux2_pos) node[yshift=0.02cm, xshift=0.15cm] (mux2_pin0) {}
(mux2_pos) node[yshift=0.02cm, xshift=-0.15cm] (mux2_pin1) {}
(mux2_1) node[yshift=0.55cm] (init5) {0}
(mux2_0) node[yshift=0.55cm] (init4) {0}
(init5) -- (mux2_pin1)
(init4) -- (mux2_pin0)

(mux2) node[xshift=1cm] (mux3_pos) {} 
(mux3_pos) node[multiplexer, xscale=0.6, yscale=0.5, rotate=-90](mux3){}
(mux3_pos) node[xshift=0.15cm] (mux3_0) {0}
(mux3_pos) node[xshift=-0.15cm] (mux3_1) {1}
(mux3_pos) node[yshift=0.02cm, xshift=0.15cm] (mux3_pin0) {}
(mux3_pos) node[yshift=0.02cm, xshift=-0.15cm] (mux3_pin1) {}
(mux3_1) node[yshift=0.55cm] (init3) {1}
(mux3_0) node[yshift=0.55cm] (init2) {0}
(init3) -- (mux3_pin1)
(init2) -- (mux3_pin0)

(mux3) node[xshift=1cm] (mux4_pos) {} 
(mux4_pos) node[multiplexer, xscale=0.6, yscale=0.5, rotate=-90](mux4){}
(mux4_pos) node[xshift=0.15cm] (mux4_0) {0}
(mux4_pos) node[xshift=-0.15cm] (mux4_1) {1}
(mux4_pos) node[yshift=0.02cm, xshift=0.15cm] (mux4_pin0) {}
(mux4_pos) node[yshift=0.02cm, xshift=-0.15cm] (mux4_pin1) {}
(mux4_1) node[yshift=0.55cm] (init1) {1}
(mux4_0) node[yshift=0.55cm] (init0) {0}
(init1) -- (mux4_pin1)
(init0) -- (mux4_pin0)

(-1,-0.6) -- (2.5,-0.6)
(-0.8,-0.6) node[yshift=0.15cm] {I1} 
(0,0) node[xshift=0.5cm, yshift=-0.6cm] (mux5_pos) {} 
(mux5_pos) node[multiplexer, xscale=0.6, yscale=0.5, rotate=-90](mux5){}
(mux5_pos) node[xshift=0.15cm] (mux5_0) {0}
(mux5_pos) node[xshift=-0.15cm] (mux5_1) {1}
(mux5_pos) node[yshift=0.02cm, xshift=0.15cm] (mux5_pin0) {}
(mux5_pos) node[yshift=0.02cm, xshift=-0.15cm] (mux5_pin1) {}
(0,0) --++(0,-0.3cm) -| (mux5_pin1)
(mux2_pos) --++(0,-0.3cm) -| (mux5_pin0)

(mux3_pos) node[xshift=0.5cm, yshift=-0.6cm] (mux6_pos) {} 
(mux6_pos) node[multiplexer, xscale=0.6, yscale=0.5, rotate=-90](mux6){}
(mux6_pos) node[xshift=0.15cm] (mux6_0) {0}
(mux6_pos) node[xshift=-0.15cm] (mux6_1) {1}
(mux6_pos) node[yshift=0.02cm, xshift=0.15cm] (mux6_pin0) {}
(mux6_pos) node[yshift=0.02cm, xshift=-0.15cm] (mux6_pin1) {}
(mux3_pos) --++(0,-0.3cm) -| (mux6_pin1)
(mux4_pos) --++(0,-0.3cm) -| (mux6_pin0)

(-1,-1.2) -- (1.5,-1.2)
(-0.8,-1.2) node[yshift=0.15cm] {I2}

(mux5_pos) node[xshift=1cm, yshift=-0.6cm] (mux7_pos) {} 
(mux7_pos) node[multiplexer, xscale=0.6, yscale=0.5, rotate=-90](mux7){}
(mux7_pos) node[xshift=0.15cm] (mux7_0) {0}
(mux7_pos) node[xshift=-0.15cm] (mux7_1) {1}
(mux7_pos) node[yshift=0.02cm, xshift=0.15cm] (mux7_pin0) {}
(mux7_pos) node[yshift=0.02cm, xshift=-0.15cm] (mux7_pin1) {}
(mux5_pos) --++(0,-0.3cm) -| (mux7_pin1)
(mux6_pos) --++(0,-0.3cm) -| (mux7_pin0)
(mux7_pos) --++ (0,-0.5)
(mux7_0) node[yshift=-0.4cm, xshift=0.3cm] () {output}
;

\draw[dashed, rounded corners, blue] (-1.8,-1.7) rectangle ++ (5.3,2.5) node[] {};
\draw[draw=none, fill=blue!50!white, opacity=0.6] (2.7,-1.5) rectangle ++(0.7,0.5) node[pos=.5] () {\textbf{LUT}};

\draw
(-2.2,0) -- (-1,0)
(-2.2,-0.6) -- (-1,-0.6)
(-2,-1.2) -- (-1,-1.2)
;

\draw[rounded corners, mypink, pattern=north east lines, pattern color=mypink] (-2.1, -0.8) rectangle ++ (0.6,1) node[] {};

\node [draw, shape=rectangle, minimum width=0.5cm, minimum height=0.2cm, fill=red, anchor=center] at (-1.8, 0) {};
\node [draw, shape=rectangle, minimum width=0.2cm, minimum height=0.2cm, fill=green!70!gray, anchor=center] at (-1.8, -0.6) {};
\node [draw=none, shape=rectangle, minimum width=0.3cm, minimum height=0.2cm, fill=white, anchor=center] at (-1.8, -1.2) {select};

\draw[thick, blue] (-2.2,-1.8) rectangle ++ (5.8,2.7) node[] {};

\draw[draw=none, fill=mypink!50!white, opacity=0.6] (-6,0.9) rectangle ++(3.4,0.5) node[pos=.5] () {\textbf{Configurable delay skew}};
\draw[thick, mypink] (-6.9,0.9) rectangle ++ (4.3,-2.8) node[] {};

\draw[mypink, dashed]
(-2.6, 1.4) -- (-2.1, 0.2)
(-2.6, -1.8) -- (-2.1, -0.8)
;
\end{scope}

\begin{scope}[xshift=-1cm,yshift=-5.55cm, yscale=0.55, xscale=0.54]
	\begin{axis}[
	ybar,
	height=4.6cm,		
	width=6.5cm,
	axis x line*=bottom, 
	axis y line*=left,
	grid=major,
	grid style={dashed,gray!50},
	ylabel = {Minimal net delay (ps)},
	xlabel = {Physical pin},
	clip=false,
	nodes near coords,
	every node near coord/.append style={font=\large},
	title = {\Large{Physical pin assignment for logical pin I0/I1}},
	xtick={1,2,3,4,5,6},
	xticklabels={A1,A2,A3,A4,A5,A6},
	ylabel style={yshift=0.3cm,xshift=-0.5cm},
        xlabel style={yshift=-0.2cm},
	title style={xshift=-1.2cm},
        label style={font=\Large},
        tick label style={font=\Large}
	]
	\addplot[
	style={black, fill=mypink!70!black}
	] 
	table[x=A, y=delay] {\PinAssign};
\end{axis}

\end{scope}

\end{circuitikz}

\caption{Time-domain popcount and comparison, implemented by PDL and arbiter, respectively.}
\label{fig:pdl}
\vspace{-0.2cm}
\end{figure}

\subsubsection{\textbf{Overall operational mechanism}}
The core concept behind the time-domain popcount and comparison design is as follows. Each PDL functions as a converter, transforming a binary code (input vector) into a cumulative delay (popcount result) based on its corresponding Hamming weight.

As illustrated in Fig. \ref{fig:pdl}, when comparing the popcount of two binary codes, these codes are represented by the signals S\_{up} and S\_{lo}, corresponding to the upper and lower PDLs, respectively. Each bit in these codes is processed using two elementary delay units: one long (high-latency net) and one short (low-latency net). The delay path selects either the longer or shorter delay unit via a multiplexer: a bit of S\_{up}/S\_{lo} equal to ``0''/``1'' inserts the longer/shorter delay unit.

Once all input binary codes are valid and ready for conversion, a start signal propagates from the left to the right end of all PDLs. The delay incurred by each PDL is inversely proportional to the Hamming weight of its corresponding binary code. In other words, a binary code with a higher popcount reaches the end earlier than one with a lower popcount. The arrival times at the right ends of the two PDLs are captured by an arbiter, implementing an argmax operation.

Specifically, in the case of TMs, the input binary code for a PDL is derived from the outputs of all clauses belonging to a particular class. As shown in Fig. \ref{fig:tm_overview} (a), half of the clauses vote for the class, while the other half vote against it. To handle this polarity within a single PDL, an input bit from a clause supporting the class (positive clause) selects the longer/shorter delay unit if it is ``0''/``1'', whereas for a clause opposing the class (negative clause), the selection is reversed: a ``1''/``0'' input inserts the longer/shorter delay unit.

\subsubsection{\textbf{PDL}} \label{sec:pdl}
Time-domain popcount is implemented using PDLs consisting of cascaded delay elements, each realized with a single LUT, similar to \cite{majzoobi2010fpga, mahalat2019efficient, anandakumar2022implementation, sahoo2015towards}. However, our design places a strong emphasis on ensuring structurally symmetric PDLs and physically identical delay elements.

Across all PDLs, the start signal—triggered by a rising or falling transition—is synchronized using D flip-flops (FFs) running at the maximum clock frequency.
This synchronization is essential, as the input transition may be distributed across a large number of PDLs, which otherwise leads to uneven signal propagation. The potential skew caused by fan-out is mitigated by allowing the transition to propagate only at the clock edge, which is uniformly distributed to all FFs through clock tree synthesis.

Each delay element is implemented by configuring a LUT to function as a multiplexer with two inputs, both connected to the output of the preceding logic. These two inputs have different propagation delays, realized by routing them through high-latency and low-latency nets, respectively, as described in Section \ref{sec:flow}. We emphasize that a specific logical pin mapping process is required to assign the inputs to physical pins, particularly for Xilinx FPGAs, where pins A6 and A5 are faster than the others, as reported in \cite{vivado}. To validate this, we evaluate the minimal net delay for all physical pins using Vivado, as shown in Fig. \ref{fig:pdl}, to determine the optimal pinout selection: the low-latency and high-latency nets are assigned to the fastest and second-fastest physical pins, respectively. The delay of the high-latency net is then adjusted during the routing phase to minimize the delay difference relative to the low-latency net, achieving minimal overall latency while ensuring adequate granularity and resolution for the task.


\subsubsection{\textbf{Arbiter}}
A NAND SR latch is employed as the arbiter to respond to the race between two PDLs, outputting ``0" or ``1" based on which chain introduces the rising transition first. The latch, constructed from two cross-coupled NAND gates, ensures symmetric placement relative to the two PDLs. An OR gate generates a completion signal to indicate the comparison is complete. For comparisons involving more than two PDLs, additional levels of arbiters are added, with the completion signal from the previous level serving as input to the next. For falling transitions, a separate arbiter is used, comprising a NOR SR latch and an AND gate to produce the comparison result and completion signal, respectively.

Metastability may occur if two PDLs trigger output transitions at nearly the same time. However, this can usually be resolved by increasing the delay difference between high- and low-latency nets of all delay elements, which improves resolution and ensures a sufficient gap between the transition arrival times, even when the two PDLs receive a nearly identical (but not the same) ones from their corresponding input vectors. Therefore, metastability may only occur if two PDLs receive equal number of ones (same Hamming weight). For certain ML operations like argmax, where two inputs are identical, the argmax function is designed to either arbitrarily select one input or consistently return a specific index. In both cases, this might be interpreted as an incorrect decision, as the result would essentially be based on a random or predetermined guess (basically, the result of inference may not match the class label in the training set)\footnote{The discussion of the interpretation of the `classification metastability' is outside the scope of this paper. It is worth noting that the problem of non-unique classification using argmax is sometimes mitigated by the techniques such as Softmax and Softermax~\cite{softmax}.}. 

\subsection{Implementation} \label{sec:flow}
We present the FPGA implementation flow for the time-domain popcount design in Fig. \ref{fig:flow}. While some steps require manual intervention, these can be performed using the example Tcl scripts provided as references. Each step is repeated for every delay element.

\definecolor{comments}{rgb}{0.5,0.5,0.5}
\definecolor{keywords}{rgb}{1, 0, 1}

\begin{figure}[!htb]
\centering
\begin{tikzpicture}[scale=0.8, every node/.style={scale=0.8}]

\begin{scope}[xshift=0cm,yshift=0cm]

\draw[draw=none, sharp corners,fill=mypink] (1.5,0.2) -- (1.6,-0.2) -- (-1.3,-0.2)   -- (-1.3,0.2) [rounded corners=10pt] -- cycle;
\draw
(0,0) node[] () {\textbf{\textcolor{white}{Placement}}}
;

\node [shape=rectangle, align=left, font=\small](placement) at (2.8,-0.8){
\begin{tabular}{l}
\textit{\textcolor{comments}{\# Place delay element in specified site and slice}} \\
\textcolor{keywords}{set\_property} BEL D6LUT [\textcolor{keywords}{get\_cells} PDL0\_0/MUX] \\
\textcolor{keywords}{set\_property} LOC SLICE\_X74Y0 [\textcolor{keywords}{get\_cells} PDL0\_0/MUX]
\end{tabular}
};
\draw[] (-1.29,-0.2) rectangle ++ (10.7,-1.2) node[] {};

\draw[draw, sharp corners] (6.5,0.2) -- (6.4,-0.2) -- (9.41,-0.2)   -- (9.41,0.2) [rounded corners=8pt] -- cycle;
\draw (8,0) node[] () {\textbf{Example scripts}};
\draw[white,line width=0.5pt] (6.41,-0.2) -- (9.4,-0.2);

\draw[-Stealth] (4.06, -1.4) -- (4.06, -2.2);
\end{scope}

\begin{scope}[xshift=0cm,yshift=-2cm]
\draw[draw=none, sharp corners,fill=mypink] (1.5,0.2) -- (1.6,-0.2) -- (-1.3,-0.2)   -- (-1.3,0.2) [rounded corners=10pt] -- cycle;
\draw
(0,0) node[] () {\textbf{\textcolor{white}{Pin Assignment}}}
;
\node [shape=rectangle, align=left, font=\small](pin) at (3.3,-0.65){
\begin{tabular}{l}
\textit{\textit{\textcolor{comments}{\# Map logical pins to physical pins}}} \\
\textcolor{keywords}{set\_property} LOCK\_PINS \{I1:A6 I0:A5\} [\textcolor{keywords}{get\_cells} PDL0\_0/MUX]
\end{tabular}
};
\draw[] (-1.29,-0.2) rectangle ++ (10.7,-0.9) node[] {};

\draw[draw, sharp corners] (6.5,0.2) -- (6.4,-0.2) -- (9.41,-0.2)   -- (9.41,0.2) [rounded corners=8pt] -- cycle;
\draw (8,0) node[] () {\textbf{Example scripts}};
\draw[white,line width=0.5pt] (6.41,-0.2) -- (9.4,-0.2);

\draw[-Stealth] (4.06, -1.1) -- (4.06, -1.9);
\end{scope}

\begin{scope}[xshift=0cm,yshift=-3.7cm]
\draw[draw=none, sharp corners,fill=mypink] (1.5,0.2) -- (1.6,-0.2) -- (-1.3,-0.2)   -- (-1.3,0.2) [rounded corners=10pt] -- cycle;
\draw
(0,0) node[] () {\textbf{\textcolor{white}{Routing}}}
;
\node [shape=rectangle, align=left, font=\small](routing) at (4.15,-1.35){
\begin{tabular}{l}
route\_design -unroute \textit{\textcolor{comments}{\# Remove existing routing}} \\
\textit{\textcolor{comments}{\# Delay-driven routing for consistent delay skew}} \\
route\_design -pins [\textcolor{keywords}{get\_pins} PDL0\_0/MUX/I1] -max\_delay 500 \\
route\_design -pins [\textcolor{keywords}{get\_pins} PDL0\_0/MUX/I0] -max\_delay 800 -min\_delay 700 \\
route\_design -preserve \textit{\textcolor{comments}{\# Preserve existing routing; route the rest}} \\
\textcolor{keywords}{set\_property} is\_route fixed 1 [\textcolor{keywords}{get\_nets} {PDL0\_0/I1 PDL0\_0/I0}] \textit{\textcolor{comments}{\# Fix routing}}
\end{tabular}
};
\draw[] (-1.29,-0.2) rectangle ++ (10.7,-2.3) node[] {};

\draw[draw, sharp corners] (6.5,0.2) -- (6.4,-0.2) -- (9.41,-0.2)   -- (9.41,0.2) [rounded corners=8pt] -- cycle;
\draw (8,0) node[] () {\textbf{Example scripts}};
\draw[white,line width=0.5pt] (6.41,-0.2) -- (9.4,-0.2);
\end{scope}

\end{tikzpicture}

\caption{Implementation flow for time-domain popcount with example Xilinx Tcl scripts.}
\label{fig:flow}
\end{figure}
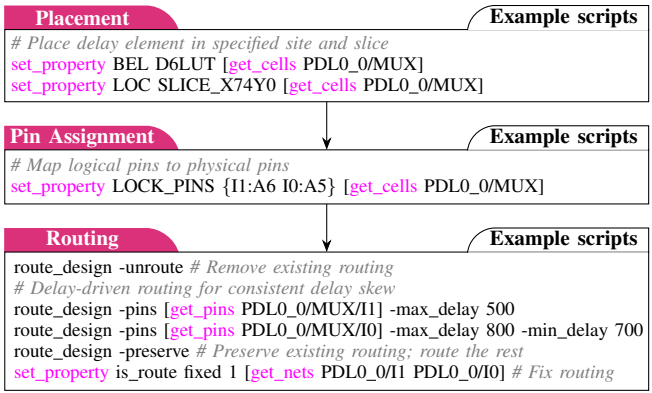

\subsubsection{\textbf{Placement}}
FPGAs consist of numerous identical logic components, which we utilize to implement uniform PDLs and delay elements. Symmetric PDLs are achieved by mapping them onto identical geometric components. Fig. \ref{fig:implementation} illustrates an example of placement on Xilinx FPGA, where PDLs are aligned vertically, with each delay element assigned to a configurable logic block (CLB). Alternative geometric placements are also possible, as long as the symmetry of the PDLs is preserved.
Similarly, the cross-coupled NAND gates in an arbiter must be symmetrically positioned relative to the corresponding PDLs.
This placement strategy increases the likelihood of achieving identical routing in later design stages. Furthermore, two cascaded delay elements are placed in adjacent CLBs, minimizing the geometric distance between them to reduce net delay for their interconnections.

\input{implementation}

\subsubsection{\textbf{Pin assignment}}
As explained in Section \ref{sec:circuit} 2), the inputs with low- and high-latency nets of a delay element are initially mapped to the fastest and second-fastest physical pins of a LUT, respectively. This minimizes overall latency and ensures sufficient delay resolution during the routing process.


\subsubsection{\textbf{Routing}}
For each delay element, we route the low- and high-latency nets by specifying the delay range, as shown in Fig. \ref{fig:flow}. With all delay elements placed at identical geometric positions within their respective CLBs and cascaded delay elements aligned relative to each other, applying the same delay ranges ensures symmetric routing across all PDLs and uniform routing for all delay elements within each PDL. This is demonstrated in Fig. \ref{fig:layout}, captured from Vivado device view for two implemented PDLs, with routing paths highlighted.
For each arbiter's two NAND gates, we specify identical physical pins, and apply the same delay constraints for routing. Similarly, we impose equal delay constraints on both inputs of each NAND gate.

\input{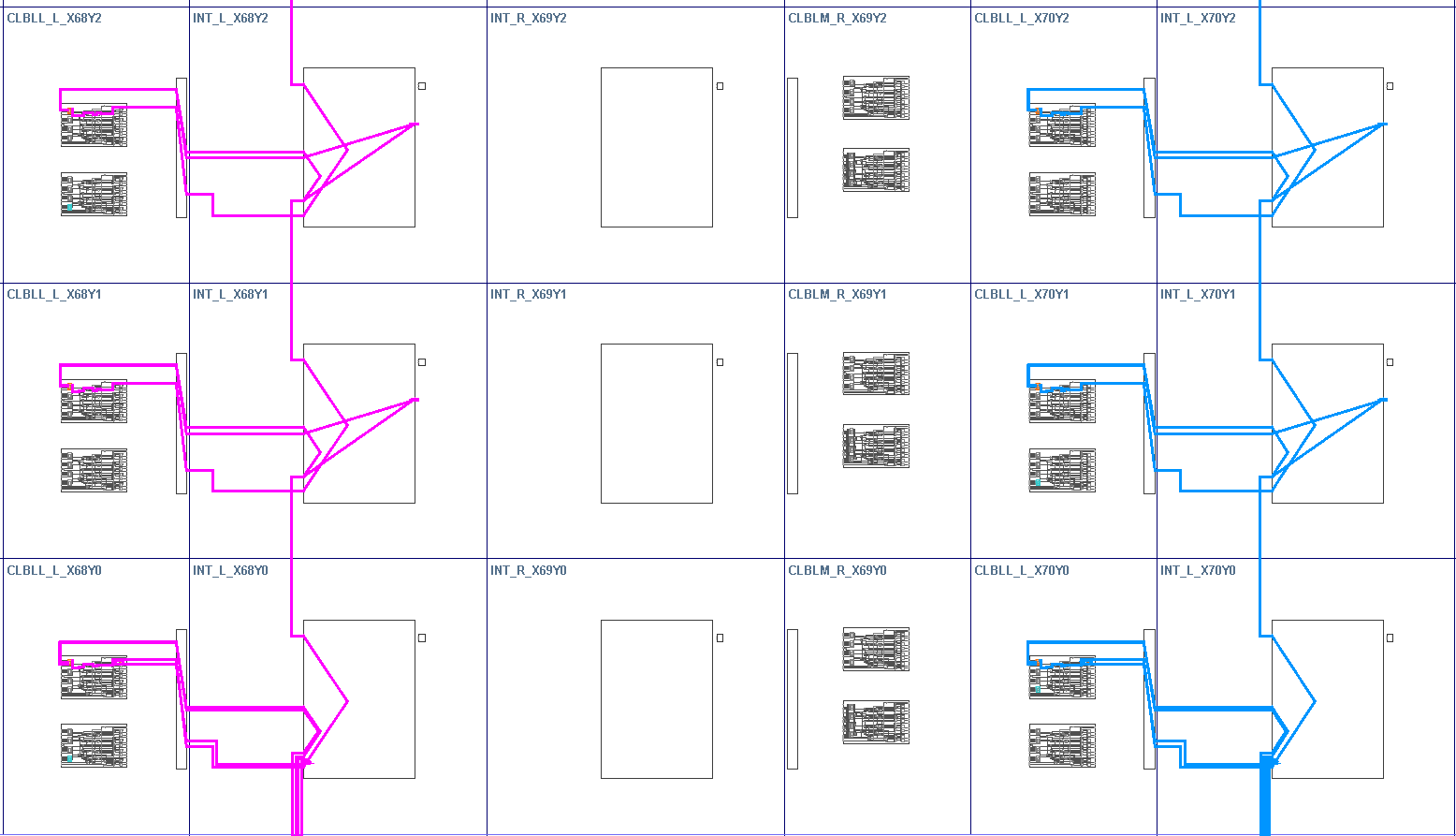}


\subsubsection{\textbf{Evaluation of Hamming weight response}}
An ideal time-domain popcount requires a monotonically decreasing propagation delay through the PDL as the input Hamming weight increases. We assess the likelihood of achieving this monotonic behavior in a practical implementation, considering the impact of process and environmental variations.

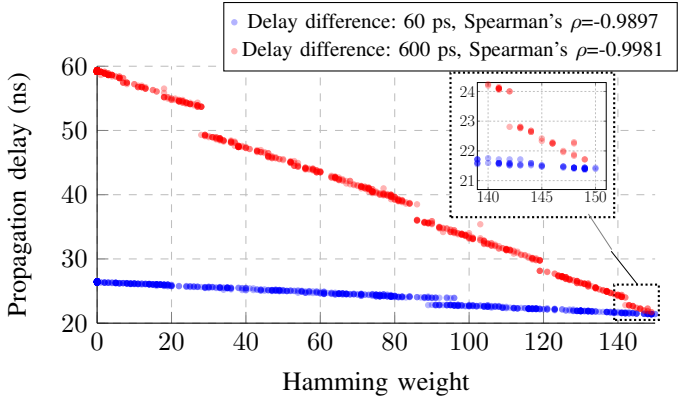
\begin{figure}[!htb]
\centering

\begin{tikzpicture}[font=\normalsize, scale=1]
\begin{axis}[
height=5cm,		
width=9cm,
axis x line*=bottom, 
axis y line*=left,
ymax=60,
ymin=20,
xmin=0,
xmax=150,
grid=major,
grid style={dashed,gray!50},
ylabel = Propagation delay (ns),
xlabel = Hamming weight,
legend style={at={(axis cs:95,70)},anchor=north,font=\footnotesize},
clip=false,
ylabel style={yshift=-0.2cm}
]	
\addplot [only marks, blue, opacity=0.3, mark size=1pt] table [y=delay, x=HD] {ila_delays_60ps.dat};
\addlegendentry{Delay difference: 60 ps, Spearman's $\rho$=-0.9897}
\addplot [only marks, red, opacity=0.3, mark size=1pt] table [y=delay, x=HD] {ila_delays_600ps.dat};
\addlegendentry{Delay difference: 600 ps, Spearman's $\rho$=-0.9981}
\coordinate[] (pt) at (axis cs:140,30);
\draw[fill=white,thick, densely dotted] (axis cs:95,36.5) rectangle (axis cs:139,59);
\draw[fill=white,thick, densely dotted] (axis cs:139,20.5) rectangle (axis cs:151,26);
\draw[line width=0.1pt] (axis cs:138.2,31.2) -- (axis cs:145.3,26);		
\end{axis}

\node[pin=93:{%
\begin{tikzpicture}[baseline,trim axis left,trim axis right,scale=0.25]
\begin{axis}[
xmin=140,xmax=150,
xtick={140,145,150},
ymin=21,ymax=24,
enlargelimits,
grid=major,
grid style={dashed,gray!50},
ticklabel style = {font=\Huge}
]
\addplot [only marks, blue, opacity=0.3, mark size=4pt] table [y=delay, x=HD] {ila_delays_60ps.dat};
\addplot [only marks, red, opacity=0.3, mark size=4pt] table [y=delay, x=HD] {ila_delays_600ps.dat};
\end{axis}
\end{tikzpicture}
}] at (pt) {};

\end{tikzpicture}

\caption{PDL propagation delay vs. input Hamming weight.}
\label{fig:delays}
\end{figure}

Specifically, we implement a PDL with 150 delay elements using the proposed design flow and measure its overall propagation delay on an FPGA board. The measurement follows the delay characterization method from \cite{majzoobi2010rapid}, with varying input Hamming weights. Fig. \ref{fig:delays} presents the measured delays for two PDLs, where the delay difference between the low- and high-latency nets is set to approximately 60 ps and 600 ps.

For each case, we compute Spearman’s rank correlation coefficient (Spearman’s $\rho$), where -1/+1 indicates a perfectly decreasing/increasing monotonic function. As shown, both cases exhibit a highly linear delay reduction as the Hamming weight increases, with Spearman’s $\rho$ extremely close to -1. Furthermore, increasing the delay difference between the low- and high-latency nets further strengthens the monotonicity.

While a perfect linear relationship between delay and Hamming weight is impossible due to intra-die process, voltage and temperature variations, the time-domain popcount can maintain sufficient accuracy by appropriately tuning the delay difference, with a trade-off in overall propagation delay as needed.

\section{TM Case Study} \label{sec:case}
\subsection{Asynchronous TM with Time-Domain Popcount} \label{sec:arch}
For the time-domain popcount presented, the critical path is determined by all delay elements with their corresponding high-latency nets, meaning the worst-case delay is primarily influenced by the cumulative delay of the high-latency nets. In tasks requiring high-accuracy popcount, the high-latency net delay must be increased to mitigate metastability in the arbiter and ensure sufficient timing resolution for the PDL. This increase in delay, however, leads to increased latency and reduced throughput when the time-domain popcount is used in a synchronous design.

Fortunately, the time-domain popcount design is naturally compatible with an asynchronous handshake protocol. The input transition of a PDL can be triggered directly by a single-bit request, while the output of either the PDL or the arbiter can be used to generate an acknowledgement or a new request, enabling the next actions with minimal additional control logic. In this configuration, the overall latency depends on the specific input vectors for the PDLs, rather than being constrained by the worst-case delay.

We present a single-rail, 2-phase asynchronous architecture for TM inference, incorporating the MOUSETRAP stage circuit \cite{singh2007mousetrap} with the time-domain popcount, as shown in Fig. \ref{fig:async_tm}, for the case study. This configuration enables high-speed operation through 2-phase handshaking and strong FPGA compatibility using simple logic gates for the control. Additionally, the time-domain popcount can be integrated with other asynchronous or self-timed architectures, 
as it effectively functions as a buffer for propagating control signals, whether level-sensitive or transition-based.

\tikzset{flipflop myD/.style={flipflop,
		flipflop def={tu=\scalebox{2.99}[5]{D}, t5=\scalebox{1}[1.67]{Q}, font=\Large}}
}

\begin{figure}[!htb]
\vspace{-0.3cm}
\centering

\begin{circuitikz}[font=\small, scale=0.92, every node/.style={scale=0.92}]

\begin{scope}[xshift=0cm,yshift=0cm]

\draw[fill=gray!50!white] (-0.35,-0.1) rectangle ++(1.7,3.7);
\draw[fill=gray] (0,0) rectangle ++(0.3,1.8) node[pos=.5, rotate=90] () {\textcolor{white}{Data latch}};
\draw[fill=gray] (0,2.2) rectangle ++(0.3,0.3);
\draw[-Stealth] (0.15,2.2) -- (0.15,1.8);
\draw[densely dotted, thick] (0.15,2.2) -- (0.15,2.5);
\draw[-Stealth] (0.15,2.7) -- (0.15,2.5);
\draw []
(0.23, 3.2) node[xnor port, scale=0.55, xscale=-1, fill=gray] (xnor_gate) {}
;
\draw[] (xnor_gate.out) -| (0.15,2.5);
\draw[] (0.3,2.35) -- (1.2,2.35);
\draw[-Stealth] (1.2,2.35) |- (xnor_gate.in 2);
\draw[] (0.5,2.35) -- (0.5,2.8) -- (0.25,2.8) to [bend left=60] (0.05,2.8);
\draw[-Stealth]  (0.05,2.8) -- (-0.6,2.8);
\node[] at (-0.1,3.2) {\textit{En}};
\draw[-Stealth] (-0.6,2.35) -- (0,2.35);
\node[] at (-0.95,2.8) {\textit{done}};
\node[] at (-0.85,2.35) {\textit{req}};
\node[] at (0.85,2.05) {\scriptsize{\textcolor{black}{Bit latch}}};
\draw[very thin] (0.37,2.1) -- (0.3,2.2);
\draw (xnor_gate.in 1) node[xshift=0.7cm] (ack) {};
\draw[-Stealth] (ack) -- (xnor_gate.in 1);
\draw (ack) node[yshift=0.2cm] () {\textit{ack}};
\draw[-Stealth] (1.2,2.35) -- (1.7,2.35);
\draw[fill=gray!50!white,rounded corners,
general shadow={%
	fill=gray,
	shadow xshift= 1pt,
	shadow yshift=-1.5pt
}] (1.7,2.2) rectangle ++(0.8,0.3) node[pos=.5] () {delay};
\draw[red] (2.5,2.35) -- ++ (1.76,0); 

\end{scope}

\node[] at (-0.9,1) {\begin{tabular}[c]{@{}c@{}} Input \\ vector \end{tabular}};
\draw[thick, -Stealth] (-1.2,0.6) -- (0,0.6);
\draw[fill=white] (1,0) rectangle ++(2.1,0.4) node[pos=.5] () {Class 0 clauses};
\draw[fill=white] (1,0.6) rectangle ++(2.1,0.4) node[pos=.5] () {Class 1 clauses};
\draw[fill=white] (1,1.2) rectangle ++(2.1,0.4) node[pos=.5] () {Class 2 clauses};
\draw[thick] (0.3,0.6) -- (0.6,0.6);
\draw[thick,-Stealth] (0.6,0.6) |- (1,0.2);
\draw[thick,-Stealth] (0.6,0.6) |- (1,0.8);
\draw[thick,-Stealth] (0.6,0.6) |- (1,1.4);

\begin{scope}[xshift=-0.5cm,yshift=0.16cm]
\begin{scope}[xshift=4.2cm,yshift=1.74cm]
\draw[fill=white] (-0.35,-0.9) rectangle ++(2.75,1.2);
\draw[] (0,0) node[flipflop myD, scale=0.25, yscale=0.65, color=red](D1){};
\node [draw=blue, shape=rectangle, minimum width=0.45cm, minimum height=0.45cm, pattern=north east lines, pattern color=blue, anchor=center, rounded corners] at (0.9, 0) {};
\node [draw=blue, shape=rectangle, minimum width=0.45cm, minimum height=0.45cm, pattern=north east lines, pattern color=blue, anchor=center, rounded corners] at (2, 0) {};
\draw[]
(D1.pin 5) -- (0.675,0)
(1.125,0) -- (1.225,0)
(1.775,0) -- (1.575,0)
(2.225,0) -- (2.4,0)
;
\draw[blue]
(0.9,-0.5) -- (0.9,-0.225)
(2,-0.5) -- (2,-0.225)
;
\node [] at (1.4,0) {\textbf{...}};
\draw[thick, blue] (-0.35,-0.5) -- (2,-0.5);
\node[] at (1.8,-0.7) {\scriptsize{Select}};
\draw[very thin] (1.5,-0.7) -- (1,-0.55);
\draw[red] (0,0.225) -- ++ (0,0.23);
\draw[red,-{Stealth[length=1.3mm]}] (0,0.455) -- (0,0.3);
\end{scope}

\begin{scope}[xshift=4.48cm,yshift=1.14cm]
\draw[fill=white] (-0.35,-0.9) rectangle ++(2.75,1.2);
\draw[] (0,0) node[flipflop myD, scale=0.25, yscale=0.65, color=red](D1){};
\node [draw=blue, shape=rectangle, minimum width=0.45cm, minimum height=0.45cm, pattern=north east lines, pattern color=blue, anchor=center, rounded corners] at (0.9, 0) {};
\node [draw=blue, shape=rectangle, minimum width=0.45cm, minimum height=0.45cm, pattern=north east lines, pattern color=blue, anchor=center, rounded corners] at (2, 0) {};
\draw[]
(D1.pin 5) -- (0.675,0)
(1.125,0) -- (1.225,0)
(1.775,0) -- (1.575,0)
(2.225,0) -- (2.4,0)
;
\draw[blue]
(0.9,-0.5) -- (0.9,-0.225)
(2,-0.5) -- (2,-0.225)
;
\node [] at (1.4,0) {\textbf{...}};
\draw[thick, blue] (-0.35,-0.5) -- (2,-0.5);
\node[] at (1.8,-0.7) {\scriptsize{Select}};
\draw[very thin] (1.5,-0.7) -- (1,-0.55);
\draw[red] (0,0.225) -- ++ (0,0.83);
\draw[red,-{Stealth[length=1.3mm]}] (0,1.055) -- (0,0.3);
\end{scope}

\begin{scope}[xshift=4.76cm,yshift=0.54cm]
\draw[fill=white] (-0.35,-0.9) rectangle ++(2.75,1.2);
\draw[] (0,0) node[flipflop myD, scale=0.25, yscale=0.65, color=red](D1){};
\node [draw=blue, shape=rectangle, minimum width=0.45cm, minimum height=0.45cm, pattern=north east lines, pattern color=blue, anchor=center, rounded corners] at (0.9, 0) {};
\node [draw=blue, shape=rectangle, minimum width=0.45cm, minimum height=0.45cm, pattern=north east lines, pattern color=blue, anchor=center, rounded corners] at (2, 0) {};
\draw[]
(D1.pin 5) -- (0.675,0)
(1.125,0) -- (1.225,0)
(1.775,0) -- (1.575,0)
(2.225,0) -- (2.4,0);
;
\draw[blue]
(0.9,-0.5) -- (0.9,-0.225)
(2,-0.5) -- (2,-0.225)
;
\node [] at (1.4,0) {\textbf{...}};
\draw[thick, blue] (-0.35,-0.5) -- (2,-0.5);
\node[] at (1.8,-0.7) {\scriptsize{Select}};
\draw[very thin] (1.5,-0.7) -- (1,-0.55);
\draw[draw=none, fill=gray!70!white, opacity=0.6] (-0.3,-0.55) rectangle ++(0.65,-0.3) node[pos=.5] () {\textbf{PDL}};
\draw[red] (0,0.225) -- ++ (0,1.43);
\draw[red,-{Stealth[length=1.3mm]}] (0,1.655) -- (0,0.3);
\end{scope}
\end{scope}   	

\draw[thick, -Stealth, blue] (3.1,0.2) -- ++ (0.85,0);
\draw[thick, -Stealth, blue] (3.1,0.8) -- ++ (0.57,0);
\draw[thick, -Stealth, blue] (3.1,1.4) -- ++ (0.29,0);

\draw[-Stealth] (6.65,0.7) -- ++(0.4,0);
\draw[-Stealth] (6.37,1.3) -- ++(0.68,0);
\draw[-Stealth] (6.09,1.9) -- ++(0.96,0);

\node[] at (6.7,2.25) {\textcolor{gray}{\scriptsize{PDL output}}};
\draw[gray, thin]
(6.85,2.15) -- (6.15,1.95)
(6.85,2.15) -- (6.43,1.35)
(6.85,2.15) -- (6.71,0.75)
;

\draw[-Stealth] (6.85,-0.1) -- ++(0,0.2) -- ++(0.2,0);
\node[] at (7.25,-0.2) {\scriptsize{const 0/1}};

\begin{scope}[xshift=7.15cm,yshift=0.4cm]
\node[] at (0.35, 0.1) {\tiny{Completion}};
\node[] at (0.52, -0.1) {\tiny{Output}};
\node[] at (0.28, 0.3) {\tiny{Input\_up}};
\node[] at (0.28, -0.3) {\tiny{Input\_lo}};
\draw[-Stealth] (0.8,-0.1) -- ++(0.25,0);
\draw[] (-0.1,-0.425) rectangle ++(0.9,0.85);
\draw[-Stealth]
(0.8,0.1) -- ++(0.35,0) --++(0,2.6) --++(-0.35,0)
;
\end{scope}

\begin{scope}[xshift=7.15cm,yshift=1.6cm]
\node[] at (0.35, 0.1) {\tiny{Completion}};
\node[] at (0.52, -0.1) {\tiny{Output}};
\node[] at (0.28, 0.3) {\tiny{Input\_up}};
\node[] at (0.28, -0.3) {\tiny{Input\_lo}};
\draw[-Stealth] (0.8,-0.1) -- ++(0.25,0);
\draw[] (-0.1,-0.425) rectangle ++(0.9,0.85);
\draw[-Stealth]
(0.8,0.1) -- ++(0.25,0) --++(0,0.8) --++(-0.25,0)
;
\end{scope}

\begin{scope}[xshift=7.15cm,yshift=2.8cm]
\draw[draw=none, fill=gray!70!white, opacity=0.6] (-0.2,0.5) rectangle ++(1.1,0.3) node[pos=.5] () {\textbf{Arbiter}};	
\node[] at (0.35, 0.1) {\tiny{Completion}};
\node[] at (0.52, -0.1) {\tiny{Output}};
\node[] at (0.45, 0.3) {\tiny{Input\_up}};
\node[] at (0.45, -0.3) {\tiny{Input\_lo}};
\draw[-Stealth] (0.8,-0.1) -- ++(0.25,0);
\draw[] (-0.1,-0.425) rectangle ++(0.9,0.85);
\draw[-Stealth] (-0.1,0.1) -- ++(-1.65,0);
\node[] at (-0.9,0.3) {\textit{Completion}};
\end{scope}

\draw[] (xnor_gate.in 1) -- ++ (2,0);
\draw[fill=white] (2.1,2.7) rectangle ++(3.3,0.8) node[pos=.5] () {Asynchronous controller};

\end{circuitikz}

\caption{Asynchronous TM architecture, integrating a MOUSETRAP stage and time-domain popcount and comparison.}
\label{fig:async_tm}
\vspace{-0.2cm}
\end{figure}
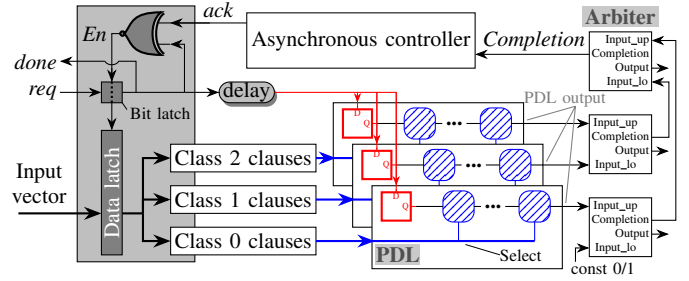

In Fig. \ref{fig:async_tm}, the processing logic following the transparent latches consists of TM clause blocks responsible for propositional logic computations, as shown in Fig. \ref{fig:tm_overview} (a). We adopt a bundled-data scheme for these clause computation blocks, requiring a bundling signal based on their worst-case delay. In FPGA, this bundling signal can be generated by appropriately setting a net delay, eliminating the need for additional logic.

Each PDL receives outputs from a TM clause block for a specific class, with the bundling signal acting as the input transition. The clause output selects whether the propagation path follows the low- or high-latency net of a delay element. As explained in Section \ref{sec:flow}, the connections of the low- and high-latency nets are swapped at the delay element inputs to account for clauses with positive and negative polarity.

Fig. \ref{fig:async_tm} illustrates a TM with three classes, requiring two levels of arbiters. At the first level, the lower arbiter has one input fixed at either 0 or 1, depending on whether $req$ undergoes a rising or falling transition in a given cycle. This ensures the arbiter is only sensitive to the incoming PDL output while maintaining a symmetric arbiter tree structure. For TMs with more classes, additional arbiter levels are required. The final classification is obtained by decoding the arbiter outputs, with the last-level arbiter generating the $Completion$ signal.

The architecture features a single MOUSETRAP stage for the present single-layer TMs. $done$ signal toggles $req$ to initiate a new inference process, enabling support for batched data. A simple asynchronous controller generates $ack$, switching the latches from opaque to transparent based on $Completion$ and all PDL outputs, as explained later in this section. Notably, this architecture can be adapted for a pipelined design with minimal modifications: $done$ serves as the acknowledgment signal for the previous stage, while the asynchronous controller generates the request signal for the next pipeline stage.

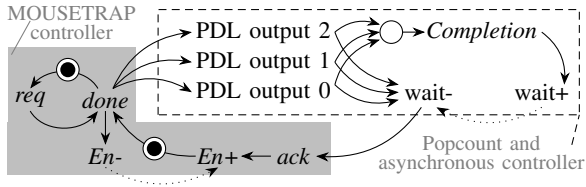
\begin{figure}[!htb]
\vspace{-0.3cm}
\centering

\begin{tikzpicture}[font=\small, every node/.style={scale=1}, scale=1]
\draw[draw=none, fill=gray!50!white] (-0.3,-1) rectangle ++(4.3,0.7) node[] () {};
\draw[draw=none, fill=gray!50!white] (-0.3,-0.3) rectangle ++(1.7,1) node[] () {};
\node[] at (0.55,1.15) {\footnotesize{\textcolor{gray}{MOUSETRAP}}};
\node[] at (0.55,0.9) {\footnotesize{\textcolor{gray}{controller}}};
\draw[thin, gray] (0,1.05) -- (-0.25,0.6);

\draw[densely dashed] (1.7,-0.2) rectangle ++(5.6,1.4) node[] () {};
\node[] at (6,-0.6) {\footnotesize{\textcolor{gray}{Popcount and}}};
\node[] at (6,-0.85) {\footnotesize{\textcolor{gray}{asynchronous controller}}};
\draw[thin, densely dashed] (7.2,-0.6) -- (7.3,-0.2);

\node[] at (0,0) {\textit{req}};
\draw[-Stealth] (0,-0.15) to [bend right=60] (0.9,-0.15);
\node[] at (1,0) () {\textit{done}};
\draw[-Stealth] (0.9,0.15) to [bend right=60] (0,0.15);
\node[draw,circle,minimum size=0.3cm,inner sep=0pt,fill=white] at (0.5,0.4) {};
\node[draw=none,circle,minimum size=0.2cm,inner sep=0pt,fill=black] at (0.5,0.4) {};

\node[] at (1,-0.75) {\textit{En-}};
\draw[-Stealth] (1,-0.15) -- (1,-0.6);

\node[] at (2.5,-0.75) {\textit{En+}};
\draw[-Stealth] (2.2,-0.75) to [bend left=30] (1.1,-0.15);
\node[draw,circle,minimum size=0.3cm,inner sep=0pt,fill=white] at (1.65,-0.65) {};
\node[draw=none,circle,minimum size=0.2cm,inner sep=0pt,fill=black] at (1.65,-0.65) {};

\draw[-Stealth, dotted] (1,-0.9) to [bend right=30] (2.5,-0.9);

\node[] at (3.5, -0.75) {\textit{ack}};
\draw[-Stealth] (3.2,-0.75) -- (2.8,-0.75);

\node[] at (3.1,0.9) {PDL output 2};
\draw[-Stealth] (1.1,0.15) to [bend left=30] (2.15,0.9);

\node[] at (3.1,0.5) {PDL output 1};
\draw[-Stealth] (1.1,0.15) to [bend left=30] (2.15,0.5);

\node[] at (3.1,0.1) {PDL output 0};
\draw[-Stealth] (1.1,0.15) to [bend left=30] (2.15,0.1);

\node[draw,circle,minimum size=0.3cm,inner sep=0pt,fill=white] at (4.8,0.9) {};
\draw[-Stealth] (4.05,0.9) to [bend left=30] (4.65,1);
\draw[-Stealth] (4.05,0.5) to [bend left=30] (4.65,0.9);
\draw[-Stealth] (4.05,0.2) to [bend left=30] (4.65,0.8);

\node[] at (5.3,0.1) {wait-};
\draw[-Stealth] (4.05,0.9) to [bend right=30] (4.9,0.2);
\draw[-Stealth] (4.05,0.5) to [bend right=30] (4.9,0.1);
\draw[-Stealth] (4.05,0.2) to [bend right=30] (4.9,0);

\node[] at (6,0.9) {\textit{Completion}};
\draw[-Stealth] (4.95,0.9) -- (5.25,0.9);

\node[] at (6.8,0.1) {wait+};
\draw[-Stealth] (6.8,0.9) to [bend left=30] (7.1,0.2);

\draw[-Stealth, dotted] (6.8,-0.1) to [bend left=30] (5.4,-0.1);
\draw[-Stealth] (5.2,-0.1) to [bend left=30] (3.8,-0.75);

\end{tikzpicture}

\caption{STG for asynchronous TM, with the dotted arc as a mandatory timing dependency not enforced by the controller.}
\label{fig:tm_stg}
\vspace{-0.3cm}
\end{figure}

We specify the overall operation using a signal transition graph (STG), shown in Fig. \ref{fig:tm_stg}. The signal transitions and their causal relationships are partially realized by the MOUSETRAP control (see \cite{singh2007mousetrap}). In our design, a merge fragment based on all PDL outputs that provides the $Completion$ signal is implemented using arbiters. A transition in the $Completion$ triggers a change in a wait signal in the asynchronous controller. This wait signal temporarily halts operations until the appropriate transitions are received at all PDL outputs, as managed by a join fragment. This suspension prevents unarrived transitions from interfering with the next inference cycle. Given that sufficient timing resolution has been achieved by appropriately setting the delay differences for the delay element inputs, the falling and rising transitions of the wait signal can always be met by the timing. Note that for each inference, the overall latency is determined by the TM producing the smallest class sum; however, in practice, it rarely reaches the worst-case scenario, where all delay elements propagate with the low-latency net, as demonstrated in Section \ref{sec:eval}.

\subsection{Experimental Setup} \label{sec:setup}
We validate the asynchronous TM using two datasets: Iris \cite{misc_iris_53} and MNIST \cite{deng2012mnist} (Table \ref{tab:dataset}). For each dataset, a Booleanization process is first applied to convert the raw features into a set of Boolean data \cite{9923830}. For Iris, each raw feature is Booleanized into three Boolean features using quantile binning, represented as a three-bit one-hot encoding, resulting in a total of 12 Boolean features. For MNIST, it is performed by applying a threshold of 75 to all grayscale values.

\begin{table}[!htb]
	\centering
	\caption{Dataset, TM model and PDL details.}
	\label{tab:dataset}

\begin{minipage}{\textwidth}
\resizebox{0.5\textwidth}{!}{
\centering
\begin{threeparttable}	
	\begin{tabular}{|p{0.5cm}p{1cm}p{1cm}|ccc|cc|}
	\hline 
	\multicolumn{3}{|c|}{\textbf{Dataset}} & \multicolumn{3}{c}{\textbf{TM}} & \multicolumn{2}{|c|}{\textbf{PDL net delay}$^a$ (ps)} \\
	 & Classes & \begin{tabular}[c]{@{}c@{}} Boolean \\ features \end{tabular} & Clauses$^b$ & ($T$,$s$) & \begin{tabular}[c]{@{}c@{}} Accuracy \\ (\%) \end{tabular} & \begin{tabular}[c]{@{}c@{}} Low- \\ latency \end{tabular} & \begin{tabular}[c]{@{}c@{}} High- \\ latency \end{tabular}  \\ \hline
	 \multirow{2}{*}{Iris} & \multirow{2}{*}{\ \ \ \ 3} & \multirow{2}{*}{\ \ \ \ 12} & \cellcolor{gray!30} 10 & \cellcolor{gray!30} (5,1.5) & \cellcolor{gray!30} 96.7 & \cellcolor{gray!30} 375.4 & \cellcolor{gray!30} 641.9 \\
	 & & & 50 & (7,6.5) & 90 & 388.6 & 593 \\ \hline
	 \multirow{2}{*}{MNIST} & \multirow{2}{*}{\ \ \ 10} & \multirow{2}{*}{\ \ \ 784} & \cellcolor{gray!30} 50 & \cellcolor{gray!30} (5,7) & \cellcolor{gray!30} 94.5 & \cellcolor{gray!30} 402.8 & \cellcolor{gray!30} 603.3 \\
	 & & & 100 & (5,10) & 95.4 & 371.1 & 632.1 \\ \hline

	\end{tabular}

	\begin{tablenotes}
	\item[$a$] Delay to realize lossless accuracy \ \ \ \ $^b$ Number of clauses per class	
	\end{tablenotes}

\end{threeparttable}
}

\end{minipage}
\end{table}

For Iris and MNIST, we train two TMs with 10 and 50 clauses, and 50 and 100 clauses, respectively, to assess performance across varying numbers of classes and clauses, for evaluation purpose. Higher accuracy could be achieved by using more clauses \cite{tarasyuk2023systematic}. The hyperparameters $T$ and $s$ for TM training are chosen based on the optimal accuracy setups from \cite{tarasyuk2023systematic}.

For PDLs, we set the low-latency net delay to the smallest possible value and adjust the high-latency net delay using trial and error to determine the minimum delay that ensures lossless accuracy. On average, the low- and high-latency net delays are 384.5 ps and 617.6 ps, respectively, with a 233.1 ps difference.

We emphasize that the proposed architecture can be applied to implement TMs for any dataset, regardless of the number of classes, clauses, or features. The trends observed in design metrics as the model scales reported in the following section are applicable to datasets beyond Iris and MNIST.

All designs were implemented on a Xilinx Zynq XC7Z020 FPGA (PYNQ-Z1), featuring 53,200 LUTs and 106,400 FFs in a 28 nm technology node.

The proposed architecture is compared to the following state-of-the-art designs:

\begin{itemize} 
\item \textbf{Generic implementation} – A synchronous TM architecture with adder-based popcount and comparison, synthesized and implemented using Vivado 2024.1’s generic process. 
\item \textbf{FPT'18} \cite{kim2018fpga} – 
A synchronous FPGA-based popcount circuit, originally designed for BNNs, which we reconstructed within a TM architecture for evaluation.
\item \textbf{ASYNC'21} \cite{wheeldon2021self} – A dual-rail asynchronous TM architecture using dual-rail 8-bit pop counters, originally presented in \cite{dalalah2006new}. 
Since this circuit is not designed for FPGA and its implementation would require extensive modifications, we compare only resource utilization by evaluating the equivalent LUT count of their pop counters, synthesizing their building blocks in Vivado.
\end{itemize}

\subsection{Evaluation and Comparison} \label{sec:eval}
We present the inference latency, resource utilization (total LUTs and FFs), and dynamic power for the evaluated implementations in Fig. \ref{fig:tm_results}. For resource utilization, we treat LUTs and FFs equally for simplicity, although in practice, their impact on area can vary depending on the design and FPGA architecture. For the two adder-based synchronous implementations, latency, defined as the minimal clock period, is determined by the worst-case critical path delay. For the proposed asynchronous design, latency is measured as the average inference time over 100 samples, as it is not controlled by clock. Resource utilization and dynamic power are obtained from the Vivado implementation reports.  

For each metric, we highlight the proportional contribution of the popcount and comparison operations. As shown in Fig. \ref{fig:tm_results} (a), in all cases, the latency due to popcount and comparison dominates the overall inference latency, with the proportion contributed by these operations increasing significantly as the model scales with more classes or clauses. These operations also result in substantial overhead in terms of resource and power, especially for small TM models like those for Iris. These findings indicate that the popcount and comparison operations are a bottleneck in TM implementations.

\input{tm_results}

\subsubsection{\textbf{Latency}}
Fig. \ref{fig:tm_results} (a) shows that while the asynchronous TM with time-domain popcount has higher latency for Iris, it outperforms adder-based implementations in larger models (especially the MNIST 50 clauses case), reducing overall inference latency by up to 38\%.


To explain this trend, we further analyze the impact of TM model scaling on the total latency of popcount and comparison, as shown in Fig. \ref{fig:trend}, providing a general perspective rather than focusing on a specific dataset or model. Specifically, the latency is influenced by the number of clauses and the number of classes, which primarily determine the proportional latency contributions of popcount and comparison, respectively.

\textit{Impact of the number of clauses on popcount latency:} In Fig. \ref{fig:trend} (a), the popcount latency-and consequently, the total latency, since the comparison latency remains constant for a fixed number of classes-follows a logarithmic increase with more clauses in the generic adder-based design, as the depth of the adder tree grows logarithmically with the input length. For the time-domain popcount, latency increases linearly with the number of clauses, since the PDL length grows proportionally. The worst case assumes all delay elements select the high-latency net, while the average case is estimated using 1,000 MNIST samples. The ±3$\sigma$ interval in the average case suggests that reaching the worst-case latency is highly improbable, especially for larger TM models.

\definecolor{myyellow}{rgb}{1,0.68,0}

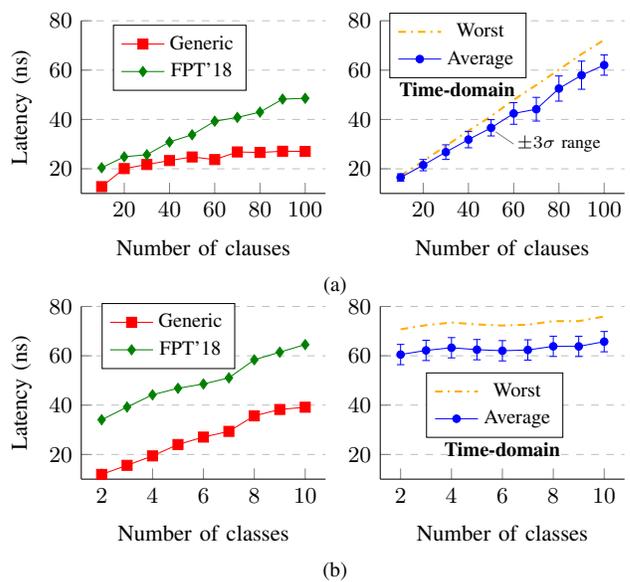
\begin{figure}[!htb]
\centering
\subfloat[]{
\begin{minipage}[t]{.24\textwidth}
\begin{tikzpicture}[font=\small, scale=0.95]		
\begin{axis}[
height=4cm,	
width=5cm,
axis x line*=bottom,
axis y line*=left,
ymax=80,
ymin=10,
xtick={20,40,60,80,100},
ymajorgrids=true,
grid style={dashed,gray!50},
ylabel = Latency (ns),
xlabel = Number of clauses,
ylabel style={yshift=-2.5ex},
legend style={at={(axis cs:15,65)},anchor=west,
font=\footnotesize}
]	
\addplot [red, mark=square*] table [y=generic, x=clauses] {TM_latency_clause.dat};
\addlegendentry{Generic}
\addplot [green!50!black, mark=diamond*] table [y=FPT, x=clauses] {TM_latency_clause.dat};
\addlegendentry{FPT'18}	
\end{axis}
\end{tikzpicture}
\end{minipage}\hspace{0cm}
\begin{minipage}[t]{.24\textwidth}
\begin{tikzpicture}[font=\small, scale=0.95]		
\begin{axis}[
height=4cm,		
width=5cm,
axis x line*=bottom,
axis y line*=left,
ymax=80,
ymin=10,
xtick={20,40,60,80,100},
ymajorgrids=true,
grid style={dashed,gray!50},
xlabel = Number of clauses,
ylabel style={yshift=-2.5ex},
legend style={at={(axis cs:5,70)},anchor=west,
font=\footnotesize},
clip=false]	
\addplot [thick, dashdotted, myyellow] table [y=proposed_worst, x=clauses] {TM_latency_clause.dat};
\addlegendentry{Worst}
\addplot [blue, mark=*, mark size=1.5pt, error bars/.cd, x dir=both, x explicit, y dir=both, y explicit] table [y=proposed_actual, x=clauses, y error=ThreeSigma] {TM_latency_clause.dat};
\addlegendentry{Average}
\node[font=\footnotesize] at (axis cs:35,52) {\textbf{Time-domain}};
\node[font=\scriptsize] at (axis cs:80,30) {$\pm3\sigma$ range};
\draw[very thin, gray] (axis cs: 60,30) -- (axis cs:52,35);
\end{axis}
\end{tikzpicture}
\end{minipage}
}
\vspace{-0.1cm}
\subfloat[]{
	\begin{minipage}[t]{.24\textwidth}
		\begin{tikzpicture}[font=\small, scale=0.95]		
		\begin{axis}[
		height=4cm,	
		width=5cm,
		axis x line*=bottom,
		axis y line*=left,
		ymax=80,
		ymin=10,
		xtick={2,4,6,8,10},
		ymajorgrids=true,
		grid style={dashed,gray!50},
		ylabel = Latency (ns),
		xlabel = Number of classes,
		ylabel style={yshift=-2.5ex},
		legend style={at={(axis cs:2,68)},anchor=west,
			font=\footnotesize}
		]	
		\addplot [red, mark=square*] table [y=generic, x=class] {TM_latency_class.dat};
		\addlegendentry{Generic}
		\addplot [green!50!black, mark=diamond*] table [y=FPT, x=class] {TM_latency_class.dat};
		\addlegendentry{FPT'18}	
		\end{axis}
		\end{tikzpicture}
	\end{minipage}\hspace{0cm}
	\begin{minipage}[t]{.24\textwidth}
		\begin{tikzpicture}[font=\small, scale=0.95]		
		\begin{axis}[
		height=4cm,		
		width=5cm,
		axis x line*=bottom,
		axis y line*=left,
		ymax=80,
		ymin=10,
		xtick={2,4,6,8,10},
		ymajorgrids=true,
		grid style={dashed,gray!50},
		xlabel = Number of classes,
		ylabel style={yshift=-2.5ex},
		legend style={at={(axis cs:3,40)},anchor=west,
			font=\footnotesize},
		clip=false]	
		\addplot [thick, dashdotted, myyellow] table [y=proposed_worst, x=class] {TM_latency_class.dat};
		\addlegendentry{Worst}
		\addplot [blue, mark=*, mark size=1.5pt, error bars/.cd, x dir=both, x explicit, y dir=both, y explicit] table [y=proposed_actual, x=class, y error=ThreeSigma] {TM_latency_class.dat};
		\addlegendentry{Average}
		\node[font=\footnotesize] at (axis cs:6,22) {\textbf{Time-domain}};
		\end{axis}
		\end{tikzpicture}
	\end{minipage}
}

\caption{Latency vs. (a) clauses (6 classes) and (b) classes (100 clauses) across different popcount implementations.}
\label{fig:trend}
\end{figure}

For FPT'18, latency also scales linearly with the number of clauses due to its ripple-carry adder-like structure, where the critical path length is determined by the input size. The increase is slightly smaller than that of the time-domain popcount in the average case, as the high-latency nets in PDLs introduce some additional overhead. These trends suggest that for large input vectors, adder-based designs may have a latency advantage over the time-domain popcount.

\textit{Impact of the number of classes on comparison latency:} Fig. \ref{fig:trend} (b) shows that overall latency in adder-based designs increases linearly with the number of classes, because each class sum must be sequentially compared, increasing comparison latency, with the popcount latency unchanged. In contrast, time-domain popcount maintains nearly constant latency, with arbiters detecting transition arrival times, and delay increases remain negligible with more arbiter levels for larger comparisons.


Thus, the time-domain popcount is particularly advantageous for tasks requiring comparisons across multiple entities, such as multi-class classifications. For TMs, this suggests even greater latency reductions in tasks with more classes than MNIST, compared to adder-based implementations.
 
\subsubsection{\textbf{Resource utilization}}
According to Fig. \ref{fig:tm_results} (b), the proposed asynchronous TM consumes the least resources in all cases except for the smallest model (10-clause TM for Iris), achieving up to a 15\% reduction in overall resource utilization.

Notably, the time-domain popcount significantly reduces resource usage compared to ASYNC'21, despite both operating in asynchronous architectures. ASYNC'21’s dual-rail adder-based popcount introduces substantial overhead beyond standard adders. While its completion detection largely enhances throughput \cite{wheeldon2021self}, it comes at a high resource cost.

\begin{figure}[!htb]
\centering
\subfloat[]{
\begin{tikzpicture}[font=\small, scale=0.95]		
\begin{axis}[
height=4cm,	
width=5cm,
axis x line*=bottom,
axis y line*=left,
xtick={20,40,60,80,100},
ytick={0,250,500,750,1000},
grid=major,
grid style={dashed,gray!50},
ylabel = LUTs and FFs,
xlabel = Number of clauses,
ylabel style={yshift=-2.5ex},
legend style={at={(axis cs:0,850)},anchor=west,
font=\footnotesize, draw=none, fill=none},
clip=false
]
\addplot [blue, mark=*] table [y=proposed, x=clauses] {TM_LUT_clause.dat};
\addlegendentry{Time-domain}
\addplot [red, mark=square*] table [y=generic, x=clauses] {TM_LUT_clause.dat};
\addlegendentry{Generic}
\addplot [green!50!black, mark=diamond*] table [y=FPT, x=clauses] {TM_LUT_clause.dat};
\addlegendentry{FPT'18}	
\end{axis}
\end{tikzpicture}
}
\subfloat[]{
	\begin{tikzpicture}[font=\small, scale=0.95]		
	\begin{axis}[
	height=4cm,	
	width=5cm,
	axis x line*=bottom,
	axis y line*=left,
	grid=major,
	grid style={dashed,gray!50},
	ylabel = LUTs and FFs,
	xlabel = Number of classes,
	ylabel style={yshift=-1ex},
	legend style={at={(axis cs:1,1450)},anchor=west,
		font=\footnotesize, draw=none, fill=none},
	clip=false
	]
\addplot [blue, mark=*] table [y=proposed, x=class] {TM_LUT_class.dat};
\addlegendentry{Time-domain}
	\addplot [red, mark=square*] table [y=generic, x=class] {TM_LUT_class.dat};
	\addlegendentry{Generic}
	\addplot [green!50!black, mark=diamond*] table [y=FPT, x=class] {TM_LUT_class.dat};
	\addlegendentry{FPT'18}	
	\end{axis}
	\end{tikzpicture}
}

\caption{Resource vs. (a) clauses (6 classes) and (b) classes (100 clauses) across different popcount implementations.}
\label{fig:LUT_trend}
\end{figure}
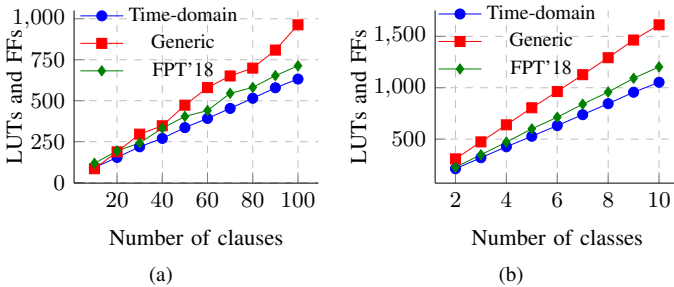

To assess whether this resource reduction scales with model size, we analyze resource usage across varying numbers of clauses and classes, as shown in Fig. \ref{fig:LUT_trend}. In all implementations, resource utilization increases linearly with model size, but the time-domain popcount consistently exhibits the smallest increment. This indicates that the resource savings achieved by the time-domain popcount is consistently maintained for larger models compared to adder-based designs. 
 
\subsubsection{\textbf{Dynamic power}}
In Fig. \ref{fig:tm_results} (c), we evaluate the dynamic power for all cases except for the 10-clause TM for Iris, as its power consumption is too small for meaningful comparison. The presented asynchronous TM achieves the lowest dynamic power consumption across both MNIST models, with reductions of up to 43.1\%. Interestingly, when comparing FPT'18 and our design for the MNIST cases, the FPT'18 popcount itself exhibits lower dynamic power than the time-domain popcount. However, the overall architecture incorporating the time-domain popcount consumes less dynamic power than the full architecture of FPT'18. This suggests that the asynchronous mechanism, by eliminating the need for a clock signal, contributes significantly to dynamic power reduction.

\begin{figure}[!htb]
\centering
\subfloat[]{
\begin{minipage}[t]{.24\textwidth}
\begin{tikzpicture}[font=\small, scale=0.95]		
\begin{axis}[
height=4cm,	
width=5cm,
axis x line*=bottom,
axis y line*=left,
xtick={20,40,60,80,100},
grid=major,
grid style={dashed,gray!50},
ylabel = Dynamic power (mW),
xlabel = Number of clauses,
ylabel style={yshift=-2.5ex},
clip=false
]	
\addplot [red, mark=square*] table [y=generic1, x=clauses] {TM_power_clause.dat};
\addplot [green!50!black, mark=diamond*] table [y=FPT1, x=clauses] {TM_power_clause.dat};
\addplot [blue, mark=*] table [y=proposed1, x=clauses] {TM_power_clause.dat};
\node[font=\footnotesize] at (axis cs:75,1.5) {\begin{tabular}[c]{@{}c@{}} \textbf{Switching activity} \\ \textbf{factor =  0.1} \end{tabular}};	
\end{axis}
\end{tikzpicture}
\end{minipage}\hspace{0cm}
\begin{minipage}[t]{.24\textwidth}
\begin{tikzpicture}[font=\small, scale=0.95]		
\begin{axis}[
height=4cm,		
width=5cm,
axis x line*=bottom,
axis y line*=left,
xtick={20,40,60,80,100},
grid=major,
grid style={dashed,gray!50},
xlabel = Number of clauses,
ylabel style={yshift=-2.5ex},
legend style={at={(axis cs:5,45)},anchor=west,fill=none,draw=none,
font=\footnotesize},
clip=false]
\addplot [blue, mark=*] table [y=proposed5, x=clauses] {TM_power_clause.dat};
\addlegendentry{Time-domain}
\addplot [red, mark=square*] table [y=generic5, x=clauses] {TM_power_clause.dat};
\addlegendentry{Generic}
\addplot [green!50!black, mark=diamond*] table [y=FPT5, x=clauses] {TM_power_clause.dat};
\addlegendentry{FPT'18}
\node[font=\footnotesize] at (axis cs:80,6) {\begin{tabular}[c]{@{}c@{}} \textbf{Switching} \\ \textbf{activity factor =  0.5} \end{tabular}};
\end{axis}
\end{tikzpicture}
\end{minipage}
}
\vspace{-0.3cm}
\subfloat[]{
	\begin{minipage}[t]{.24\textwidth}
		\begin{tikzpicture}[font=\small, scale=0.95]		
		\begin{axis}[
		height=4cm,	
		width=5cm,
		axis x line*=bottom,
		axis y line*=left,
		grid=major,
		grid style={dashed,gray!50},
		ylabel = Dynamic power (mW),
		xlabel = Number of classes,
		ylabel style={yshift=-2.5ex},
legend style={at={(axis cs:1,25)},anchor=west,fill=none,draw=none,
	font=\footnotesize}
		]
\addplot [blue, mark=*] table [y=proposed1, x=class] {TM_power_class.dat};
\addlegendentry{Time-domain}
\addplot [red, mark=square*] table [y=generic1, x=class] {TM_power_class.dat};
\addlegendentry{Generic}
\addplot [green!50!black, mark=diamond*] table [y=FPT1, x=class] {TM_power_class.dat};
\addlegendentry{FPT'18}
\node[font=\footnotesize] at (axis cs:7.5,4) {\begin{tabular}[c]{@{}c@{}} \textbf{Switching activity} \\ \textbf{factor =  0.1} \end{tabular}};
		\end{axis}
		\end{tikzpicture}
	\end{minipage}\hspace{0cm}
	\begin{minipage}[t]{.24\textwidth}
		\begin{tikzpicture}[font=\small, scale=0.95]		
		\begin{axis}[
		height=4cm,		
		width=5cm,
		axis x line*=bottom,
		axis y line*=left,
		xtick={2,4,6,8,10},
		grid=major,
		grid style={dashed,gray!50},
		xlabel = Number of classes,
		ylabel style={yshift=-2.5ex},
		clip=false]	
\addplot [red, mark=square*] table [y=generic5, x=class] {TM_power_class.dat};
\addplot [green!50!black, mark=diamond*] table [y=FPT5, x=class] {TM_power_class.dat};
\addplot [blue, mark=*] table [y=proposed5, x=class] {TM_power_class.dat}; 
\node[font=\footnotesize] at (axis cs:8.5,10) {\begin{tabular}[c]{@{}c@{}} \textbf{Switching} \\ \textbf{activity factor =  0.5} \end{tabular}};
		\end{axis}
		\end{tikzpicture}
	\end{minipage}
}

\caption{Power vs. (a) clauses (6 classes) and (b) classes (100 clauses) across different popcount implementations.}
\label{fig:power_trend} 
\vspace{-0.6cm}
\end{figure}
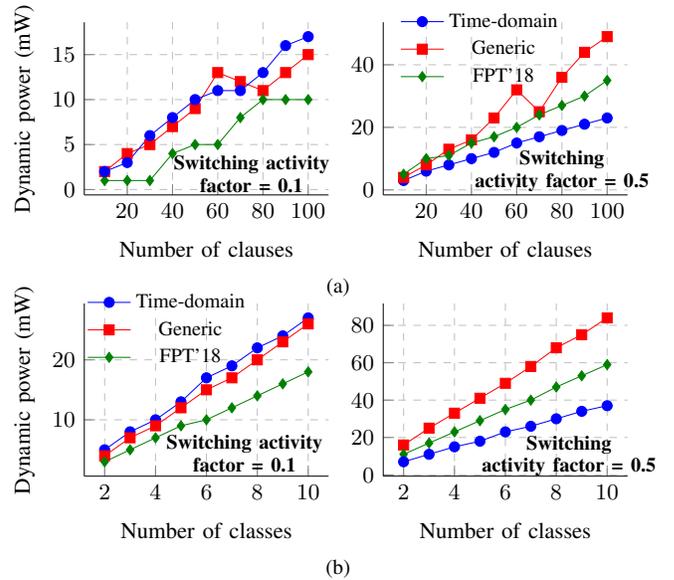

We evaluate the dynamic power of different popcount implementations while scaling the number of TM clauses and classes, as shown in Fig. \ref{fig:power_trend}. By definition, dynamic power is largely influenced by switching activity, meaning the power values reported in Fig. \ref{fig:tm_results} (c) are dependent on the dataset and input samples. To investigate this impact, we measure power consumption under two sets of input vectors, leading to switching activity factors of 0.1 and 0.5.

As shown in Fig. \ref{fig:power_trend}, for low switching activity (0.1), adder-based popcount consumes less dynamic power due to reduced circuit switching. However, in the time-domain popcount, transitions occur in all delay elements during each cycle, leading to relatively higher power consumption. On the other hand, adder-based popcount is highly sensitive to switching activity, with a significant increase in dynamic power when the switching activity factor rises to 0.5. In contrast, the time-domain popcount remains much less affected by increased switching activity, ultimately becoming the most power-efficient option.

This more stable dynamic power behavior makes the time-domain popcount advantageous in scenarios where predictable energy consumption is critical. The reduced sensitivity to switching activity simplifies power management, which is particularly beneficial for battery-powered or energy-constrained devices, where TMs and other low-complexity ML algorithms are more likely to be deployed.


\section{Conclusions and Future Work} \label{sec:conc}
We present an efficient FPGA implementation of time-domain popcount and comparison. This design leverages carefully engineered PDLs to achieve a highly linear and monotonic relationship between input Hamming weight and propagation delay. We demonstrate that the time-domain popcount can be practically implemented on an FPGA while maintaining lossless accuracy for low-complexity machine learning algorithms like TMs, serving as a case study in this paper. Exploiting the natural compatibility of the time-domain popcount with asynchronous architectures, we implement an asynchronous TM that achieves up to 38\% lower inference latency particularly for classification tasks with many classes, consistently reduces resource utilization by up to 15\%, and lowers dynamic power consumption by up to 43.1\%, while also exhibiting more stable power behavior compared to TMs using conventional adder-based popcount.

For future work, we will extend this approach to an asynchronous pipelined BNN architecture. As discussed in Section \ref{sec:arch}, the design can be adapted for pipelined architectures with minimal modifications. The output layer can follow a similar structure to Fig. \ref{fig:async_tm}, while for hidden layers, each neuron can be assigned a dedicated PDL, with inputs derived from synapse outputs computed via XNOR. Sign activation can be performed using a shared PDL with an equal number of ones and zeros as a neutral latency reference, with an arbiter determining neuron activation based on the timing relative to the neutral PDL.

\section*{Acknowledgement}
This work was supported by the Engineering and Physical Sciences Research Council (EPSRC) under Grant EP/X039943/1 and Grant EP/X036006/1.

\bibliographystyle{IEEEtran}
\bibliography{sample-base.bib}

\end{document}